\documentclass{article}

\usepackage[preprint,nonatbib]{neurips_2024}

\usepackage{biblatex}
\usepackage[utf8]{inputenc} 
\usepackage[T1]{fontenc}    
\usepackage{hyperref}       
\usepackage{url}            
\usepackage{booktabs}       
\usepackage{amsfonts}       
\usepackage{nicefrac}       
\usepackage{microtype}      
\usepackage{xcolor}         
\usepackage{graphicx}
\usepackage{caption}
\usepackage{subcaption}
\usepackage{amsmath,amsthm}
\usepackage{algorithm}
\usepackage[noend]{algpseudocode}
\def\boxit#1#2{%
    \smash{\color{blue}\fboxrule=1pt\relax\fboxsep=2pt\relax%
    \llap{\rlap{\fbox{\phantom{\rule{#1}{#2}}}}~}}\ignorespaces
}

\title{An Approximate Dynamic Programming Framework for Occlusion-Robust Multi-Object Tracking}

\author{%
  Pratyusha Musunuru\\
  Arizona State University\\
  \texttt{pmusunu1@asu.edu} \\
  \And
  Yuchao Li \\
  Arizona State University\\
  \texttt{yuchaoli@asu.edu} \\
  \And
  Jamison Weber \\
  Arizona State University\\
  \texttt{jwweber@asu.edu} \\
  \And
  Dimitri Bertsekas \\
  Arizona State University\\
  \texttt{dimitrib@mit.edu} \\
}

\begin{document}

\maketitle

\begin{abstract}
In this work, we consider data association problems involving multi-object tracking (MOT). In particular, we address the challenges arising from object occlusions. We propose a framework called approximate dynamic programming track (ADPTrack), which applies dynamic programming principles to improve an existing method called the base heuristic. Given a set of tracks and the next target frame, the base heuristic extends the tracks by matching them to the objects of this target frame directly. In contrast, ADPTrack first processes a few subsequent frames and applies the base heuristic starting from the next target frame to obtain tentative tracks. It then leverages the tentative tracks to match the objects of the target frame. This tends to reduce the occlusion-based errors and leads to an improvement over the base heuristic. When tested on the MOT17 video dataset, the proposed method demonstrates a 0.7\% improvement in the association accuracy (IDF1 metric) over a state-of-the-art method that is used as the base heuristic. It also obtains improvements with respect to all the other standard metrics. Empirically, we found that the improvements are particularly pronounced in scenarios where the video data is obtained by fixed-position cameras.   
\end{abstract}

\section{Introduction}
In this work, we consider the problem of \emph{multi-object tracking} (MOT), which can be viewed as a special case of the multidimensional assignment (MDA) problem~\cite{emami}. It involves the assignment of track identifiers to objects in motion over a sequence of image frames. In general, ensuring consistent identifiers over long sequences of frames is a formidable challenge, especially when the assignments are computed in real-time. To strike a good balance of performance and computational expedience, the so-called \emph{online tracking} methods have been researched in the literature~\cite{sort, bytetrack, botsort, cbiou}. Given a set of tracks and a new image frame, referred to as a \emph{target frame}, the online tracking methods extend the given tracks by assigning objects in the target frame to the tracks. During this process, they rely exclusively on the information of objects' movement and/or appearance in the tracks and the target frame. This makes them prone to errors in cases where objects become partially/fully occluded over one or more contiguous frames.

To overcome the occlusion challenge, we propose a method based on approximate dynamic programming (also known as reinforcement learning) techniques, which we call \emph{approximate dynamic programming track} (ADPTrack for short). It relies on an arbitrary given online tracking method and improves upon it. In particular, given a set of tracks and a target frame, ADPTrack collects a few additional subsequent frames beyond the target frame and applies the online tracking method to construct tentative tracks starting from the target frame. It then relies on the information in the given tracks and the tentative tracks for assigning objects in the target frame to the given tracks. Since future information beyond the target frame is needed, our scheme can be viewed as a \emph{near-online} method. Despite the additional computation compared with the online methods, the proposed method offers a systematic approach to address mismatches due to occlusion.

Some existing near-online methods~\cite{nomt1, nomt2} share similarities with our proposed framework. In particular, given a set of tracks, a target frame, and a few subsequent frames beyond it, they construct tentative tracks of objects in the target frame and the subsequent frames. These tentative tracks either directly extend the given tracks to the objects in the target frame, or provide information that is useful for computing weights on arcs connecting these two. Despite the similarity, these schemes have been designed based on heuristic grounds and/or require specific neural networks. In contrast, ADPTrack is a general framework that is built on the connection between MOT and dynamic programming (DP). The process of constructing tentative tracks is viewed as performing \emph{near-online simulation}. These tracks are used to approximate the \emph{optimal value function} in the context of DP. As a result, ADPTrack is flexible and can leverage arbitrary online tracking methods.

\begin{figure}[h]
    \centering
    \begin{subfigure}[b]{0.22\textwidth}
        \includegraphics[width=\textwidth]{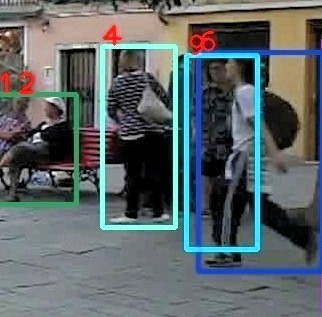}
        \caption{}
    \end{subfigure}
    \hfill
    \begin{subfigure}[b]{0.22\textwidth}
        \includegraphics[width=\textwidth]{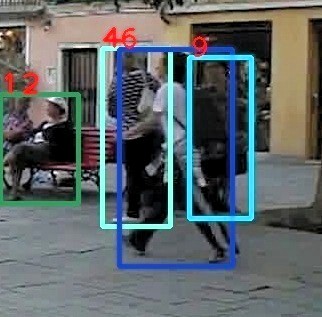}
        \caption{}
    \end{subfigure}
    \hfill
    \begin{subfigure}[b]{0.22\textwidth}
        \includegraphics[width=\textwidth]{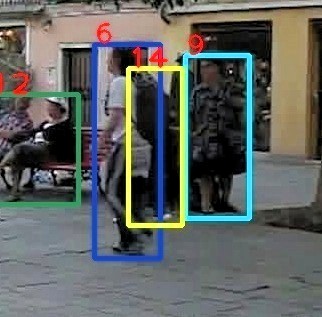}
        \caption{}
    \end{subfigure}
    \hfill
    \begin{subfigure}[b]{0.22\textwidth}
        \includegraphics[width=\textwidth]{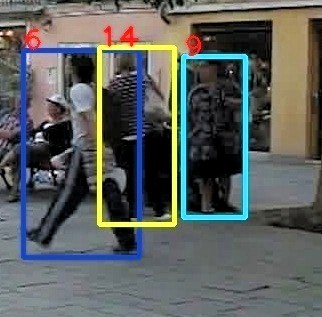}
        \caption{}
    \end{subfigure}
    \hfill
    \begin{subfigure}[b]{0.22\textwidth}
        \includegraphics[width=\textwidth]{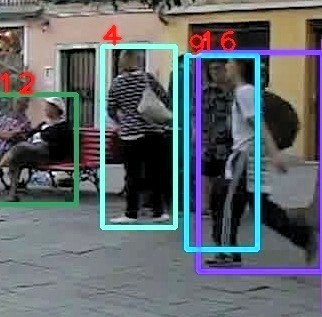}
        \caption{}
    \end{subfigure}
    \hfill
    \begin{subfigure}[b]{0.22\textwidth}
        \includegraphics[width=\textwidth]{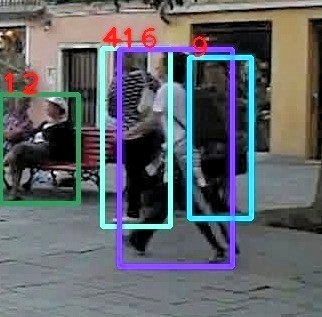}
        \caption{}
    \end{subfigure}
    \hfill
    \begin{subfigure}[b]{0.22\textwidth}
        \includegraphics[width=\textwidth]{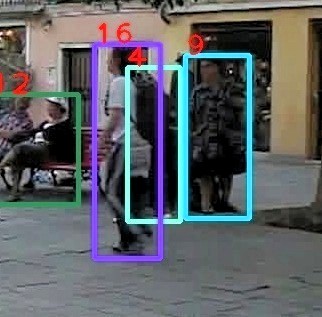}
        \caption{}
    \end{subfigure}
    \hfill
    \begin{subfigure}[b]{0.22\textwidth}
        \includegraphics[width=\textwidth]{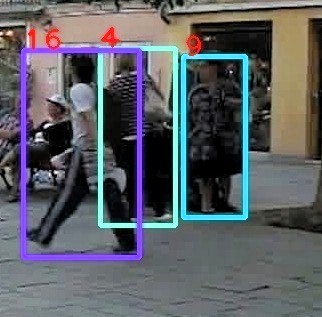}
        \caption{}
    \end{subfigure}
    \caption{Frame-by-frame comparison of BoT-SORT (a-d) versus ADPTrack with BoT-SORT as the base heuristic (e-h) in an example. Using BoT-SORT directly, person (4) in (a) is erroneously assigned to another person's identifier (14) in (c)-(d) after occluded by the person (6) in (b). When applying ADPTrack with BoT-SORT as the base heuristic, the same person (4) is assigned with the same identifier in (a) and (c)-(d) despite being occluded by the person (16) [the person (6) in (b)].}
    \label{fig:main-example}
\end{figure}

To evaluate the performance of the ADPTrack framework, we apply as the base heuristic the BoT-SORT method~\cite{botsort},  one of the state-of-the-art open-source online MOT algorithms at the time of writing. We addressed the problems in the MOT17 dataset~\cite{milan2016mot16} and obtained an overall relative improvement of 0.7\% in the IDF1 score over BoT-SORT with a minor improvement in accuracy (MOTA, HOTA metrics).
This implies ADPTrack is useful for reducing false positives, false negatives, ID switches, and fragmentation with respect to the ground truth, and is flexible to leverage any given online tracking method. Fig.~\ref{fig:main-example} provides a frame-by-frame comparison of BoT-SORT and ADPTrack with BoT-SORT as base heuristic for a fixed video example and illustrates characteristic occlusion scenarios where ADPTrack outperforms BoT-SORT.

\section{Related Work}\label{sec:related}
\textbf{Multi-Object Tracking}: We provide a brief discussion of online and near-online tracking methods that are closely related to ADPTrack. Online tracking methods perform frame-by-frame association based on the information contained in the given tracks and the target frame. 
They often involve motion~\cite{botsort, motiontrack, ocsort, strongsort} and/or appearance models~\cite{botsort, deepsort, corrtracker, p3aformer, cstrack, stam, utm, conf, poi}. 
A motion model estimates the position and size of an object in the next frame using the object's velocity. 
Typically, the motion model is a Kalman filter with the assumption that the object moves at a constant velocity~\cite{sort, bytetrack, botsort, deepsort} or at a linear combination of recent frames' velocities~\cite{cbiou}. An appearance model extracts the visual information of an object to compute similarity scores for performing the assignment.
One obtains similarity scores between objects either via deep neural networks directly~\cite{siamese1, siamese2, siamese4, dan, uma} or using a distance metric between deep appearance features~\cite{botsort, deepsort, motdt, jde, fairmot}. 
Several studies~\cite{ran, ttu, strn, siamese3, lsst, soda, dman,  transtrack, transmot, tadam,  trackformer, utt, memot, tpagt, quasidense} 
directly leverage aggregated appearance information of given tracks for assignment, indicating the importance of past information.

Near-online tracking methods perform frame-by-frame association similar to online tracking, whilst leveraging a limited number of subsequent frames~\cite{nomt1, nomt3, trackmpnn, MCMC, multiplex}.
This effectively introduces a delay for the benefit of more accurate assignments.
In~\cite{nomt2}, the authors generate tentative tracks by visual comparison and generate assignment weights using three different neural networks.
In~\cite{flow}, the authors compartmentalize the problem locally around a time window and perform assignments for the target frame using flow-based optimization techniques.
In contrast, ADPTrack builds upon existing online tracking methods and therefore does not rely on additional offline training.

\textbf{Approximation in Value Space}: 
Approximation in value space~\cite{bertsekas2023course} is a broad class of reinforcement learning methods, which have been essential in some high-profile success stories, such as AlphaGo~\cite{silver2016mastering}, AlphaZero~\cite{silver2017mastering,silver2018general}, and TD-Gammon~\cite{tesauro1996line}. 
The key idea in approximation in value space is to obtain approximations of the optimal value function in the context of DP. 
One representative method is called the \emph{rollout algorithm}, which improves upon a given base heuristic via real-time simulation and has been applied in a variety of applications, e.g., \cite{bertsekas1997rollout,bertsekas1999rollout,hu2022optimal,bhambri2022reinforcement,weber2023distributed,li2024most}. 
While our approach is not mathematically equivalent to the rollout algorithm, it uses several rollout ideas, including the simulation of a base heuristic to perform approximation in value space. 

\section{Preliminaries and Mathematical Model}\label{sec:preliminaries}
In this section, we first present an overview of the MDA problem as a generalization of MOT, its DP formulation, and its exact solution method via DP. Then we provide a mathematical description of a typical online tracking method using the DP formulation introduced earlier. 

\subsection{Multi-Object Tracking as Multi-Dimensional Assignment}\label{multidim}
An instance of the $N$-dimensional assignment problem is represented by an $(N+1)$-partite graph arranged in layers $\mathcal{N}_0,\mathcal{N}_1,\ldots,\mathcal{N}_N$, each of which contains exactly $m$ nodes. 
The arcs of the graph take the form $(i,j)$, where $i$ is a node in layer $\mathcal{N}_k$ and $j$ is a node in layer $\mathcal{N}_{k+1}$, $k=0,1,\ldots,N-1$. 
We refer to a subset of $N+1$ nodes $i_0,i_1,\ldots,i_{N}$, where $i_k\in {N}_k$ for all $k$, and their corresponding arcs $(i_0,i_1),(i_1,i_2),\ldots,(i_{N-1},i_N)$ as a \emph{grouping}. Each grouping has an associated value.
A feasible solution to an instance of the $N$-dimensional assignment problem is a set of $m$ node-disjoint groupings. 
A solution is optimal if the sum of all grouping values is maximized. When $N=1$, the problem is called \emph{bipartite matching} and can be solved exactly in polynomial time. 
We focus on the case of the MDA problem where $N>1$, which is NP-hard~\cite{MAPNP}, and hence only approximate solutions are known.

When modeling MOT as an MDA, each of the $N+1$ layers represents a frame of $m$ objects. A grouping represents an association of a single object over multiple frames and can be viewed as a (complete) track. A given track with a length smaller than $N+1$ is viewed as a partial grouping. The value of a grouping should be large if the object associated with each frame is consistent according to some ground truth. In particular, the optimal solution assigns each object to the groupings as given by the ground truth.\footnote{Note that for real-world data association problems, the number of objects in a frame need not be fixed, as objects may appear and disappear. Our theoretical model can be extended to account for these situations. For the convenience of presentation, we make the simplifying assumption that there are $m$ objects in each frame throughout our discussion. On the other hand, the implementation of our solution handles the cases where the numbers of objects vary over frames.}

\subsection{Dynamic Programming for Multidimensional Assignment Problems}

Let us now formulate the MDA problem as a deterministic DP problem. The exact solution to this DP problem is intractable. On the other hand, the DP formulation brings to bear approximation in value space, a major approach for approximate DP and reinforcement learning, which forms the basis of our proposed algorithm in Section~\ref{sec:algorithm}.

Generally, deterministic DP is used to solve a problem of sequential decision-making over $N$ stages, by breaking it down into a sequence of simpler single-stage problems. It aims to find a sequence of decisions or actions $u_0,\ldots,u_{N-1}$ by generating a corresponding sequence of optimal value functions $J^*_1,\ldots,J_{N-1}^*$. The algorithm uses a known value function $J_N^*$ to compute the next value function $J^*_{N-1}$, by solving a single-stage decision problem whose optimization variable is $u_{N-1}$. It then uses $J^*_{N-1}$ to compute $J^*_{N-2}$, and proceeds similarly to compute all the remaining value functions $J_{N-3}^*,\ldots,J^*_1$. 

More specifically, a deterministic DP problem involves a discrete-time dynamic system of the form 
 $x_{k+1}=f_k(x_k,u_k)$, $k=0,\dots,N-1$, where $k$ is a time index, $x_k$ is called the \emph{state} of the system at time $k$, which belongs to a set $X_k$, and $u_k$ is the \emph{control}, to be selected at time $k$ from a given set $U_k$. The function $f_k$ describes the mechanism by which the state is updated from $k$ to $k+1$ under the influence of a control. Given an initial state $x_0$, in our DP problem we aim to maximize the value of a given terminal function $G$, i.e., to select the sequence of controls $(u_0,\dots,u_{N-1})$ such that the value of $G(x_N)$ is maximized.  

To model the MDA problem as a DP problem, we define the set $U_k$ as the collection of all bipartite matchings between $\mathcal{N}_k$ and $\mathcal{N}_{k+1}$, i.e., each of its element $u_k$ is a set of arcs $\{(i_n,j_n)\,\mid\,n=1,\dots,m\}$ that forms a legitimate assignment between nodes in $\mathcal{N}_k$ and $\mathcal{N}_{k+1}$. The state evolution is given by $f_0(x_0,u_0)=u_0$, and $f_k(x_k,u_k)=(x_k,u_k)$ for $k=1,\dots,N-1$, and  $x_0$ is an artificial state. In particular, the state $x_k=(u_0,\dots,u_{k-1})$ is the set of $m$ partial groupings with objects the first $k+1$ layers. The value that we aim to maximize is $G(x_N)=G(u_0,\dots,u_{N-1})$. Here $x_N$ represents a feasible set of $m$ groupings (as defined in Section 3.1), and $G(x_N)$ represents the sum of the values of these $m$ groupings.

The exact DP algorithm involves computing a sequence of optimal value functions $J^*_{k}$, $k=1,\dots,N$. It first sets 
$$J^*_N(x_N)=G(x_N)=G(u_0,\ldots,u_{N-1}),$$
and then computes backwards
\begin{equation}\label{eq:bellman}
    J^*_k(x_k)=\max_{u_k\in U_k}J^*_{k+1}(x_k,u_k), \quad\text{for all }x_k,\,k=1,\dots,N-1.
\end{equation}
Having completed this calculation, it then computes the optimal controls $(u_0^*,\dots,u^*_{N-1})$ via 
\begin{equation}
\label{eq:dp_control}
    u^*_{k}\in\arg\max_{u_k\in U_k}J^*_{k+1}(x^*_k,u_k),\quad k=0,\dots,N-1,
\end{equation}
where $x^*_0$ is the artificial state, and $x_{k+1}^*=f_k(x^*_k,u_k^*)$. Although theoretically appealing, this exact method cannot be used to solve MOT due to reasons discussed later. 

\subsection{Online Tracking Methods}
Let us now provide a mathematical description of a typical online tracking method using the DP terminology introduced earlier, which we will use as a base heuristic. Given the $k$th frame, we denote by $I_k^j$ the information associated with the $j$th object in this frame that is useful for matching. In particular, we assume that $I_k^j=(v_k^j,p_k^j)$, where $v_k^j$ is the visual information associated with the $j$th object in the $k$th frame, and $p_k^j$ is the position and size information associated with the same object. They are used in the appearance models and motion models discussed in Section~\ref{sec:related}. Suppose that $m$ tracks have been formed for objects contained in the $0$th to $k$th frames via $x_k=(u_0,\dots,u_{k-1})$. We denote by $T^i(x_k)$ the information associated with the $i$th track, so that 
$$T^i(x_k)=(v_0^i,v_1^i,\dots,v_k^i,p_0^i,p_1^i,\dots,p_k^i).$$
Note that here the indices of the objects have been assigned according to the indices of the tracks that they belong to, which depends on $x_k=(u_0,\dots,u_{k-1})$. Given a track $T^i(x_k)$ and the information vector associated with the $j$th object in the $(k+1)$th frame, the base heuristic computes a weight $w_{k+1}^{ij}(x_k)$ by some function $H$, i.e., 
\begin{equation}
    \label{eq:base_w}
    w_{k+1}^{ij}(x_k)=H\big(T^i(x_k),I_{k+1}^j\big).
\end{equation}
Once the weights $w_{k+1}^{ij}(x_k)$ are computed for all track-object pairs $(i,j)$, a standard bipartite matching is performed to extend the tracks from $T^i(x_k)$ to $T^i(x_{k+1})$. From the perspective of MOT, the weights $w_{k+1}^{ij}(x_k)$ measure the ``similarity" between the first $k+1$ objects in grouping $i$ specified by $x_k$, and the object $j$ in the $(k+1)$th frame. This can be viewed as an approximation to the grouping values and leads to errors in the assignment. As we will show shortly, our proposed framework improves upon the online tracking methods by constructing better grouping value approximations.   

\section{Approximate Dynamic Programming Track}\label{sec:algorithm}
The exact and approximate methods discussed in Section~\ref{sec:preliminaries} have their respective drawbacks when used for MOT. The DP method faces two major challenges. The first applies specifically to MOT, namely that evaluating the quality of a matching sequence is difficult and thus the function $G$ is not known. Secondly, even if $G$ were known, computing $J^*_{k+1}$ is still intractable as the size of $X_{k+1}$ grows exponentially with $k$. The online tracking methods use approximate grouping values $w_{k+1}^{ij}(x_k)$ to approximate $G$ [cf.\eqref{eq:base_w}]. They allow real-time computation but also introduce assignment errors due to the simplifications introduced by the weights. {This is particularly true when the objects are occluded in the target frame.} In contrast, the proposed ADPTrack framework leverages the DP formulation and improves the quality of approximation used in online tracking methods {by considering the information contained in the subsequent frames, thus addressing the challenges posed by occlusion}. In what follows, we describe the ADPTrack as a form of approximation in value space, provide the procedure through which the approximation to the optimal value function is computed, and provide details on the implementation with BoT-SORT as the base heuristic.

\subsection{Approximation in Value Space for Multidimensional Assignment}

The approximation in value space method introduces various approximations to the components in the maximization from Equation~\eqref{eq:dp_control}. One such approximation replaces value functions $J^*_{k+1}$, $k=0,\dots,N-1$, in \eqref{eq:dp_control} with value function approximations $\tilde{J}_{k+1}$ that can be computed efficiently. To facilitate the corresponding maximization calculation, {we consider functions  $\tilde{J}_{k+1}$ of the form} 
\begin{equation}\label{eq:kplus1tilde}
    \tilde{J}_{k+1}(x_k,u_k)=\sum_{(i,j)\in u_k}c_{k+1}^{ij}(x_k)
\end{equation}
where each $(i,j)\in u_k$ is an arc from the valid perfect matching specified by control $u_k$, and $c_{k+1}^{ij}(x_k)$ represents the weight for arc $(i,j)$ where $i\in\mathcal{N}_k$ and $j\in \mathcal{N}_{k+1}$. As a result, given the current state $x_k=(\Tilde{u}_0,\dots,\Tilde{u}_{k-1})$, the control is selected via the maximization
\begin{equation}\label{eq:approxbellman}
    \Tilde{u}_k\in\arg\max_{u_k\in U_k}\tilde{J}_{k+1}(x_k,u_k),
\end{equation}
which is equivalent to solving a bipartite matching problem, with weights on arcs $(i,j)$ specified by $c_{k+1}^{ij}(x_k)$. Compared with the weight $w_{k+1}^{ij}(x_k)$ used in online methods, the weight $c_{k+1}^{ij}(x_k)$ represents the ``similarity" between the first $k+1$ objects (contained in $0$th to $k$th frames) in grouping $i$ assigned by $x_k$, and the tentative track starting from the object $j$ in the $(k+1)$th frame. Here the tentative track involves $\ell+1$ objects starting from the $(k+1)$th frame, with $\ell$ being a \emph{truncated horizon} in DP terminology. Thus we refer to $c_{k+1}^{ij}(x_k)$ as \emph{similarity scores} between tracks. An illustration of the key components involved in computing $\Tilde{J}_{k+1}$ is shown in Fig.~\ref{fig:copy-yuchao}. Intuitively, these scores contain more information than $w_{k+1}^{ij}(x_k)$, which tends to improve the assignment quality, as we discuss next.

\subsection{Value Function Approximation via {the} Base Heuristic}\label{mainalg}

\begin{figure}
\centering
\begin{subfigure}[b]{\textwidth}
    \centering
    \includegraphics[width=\textwidth]{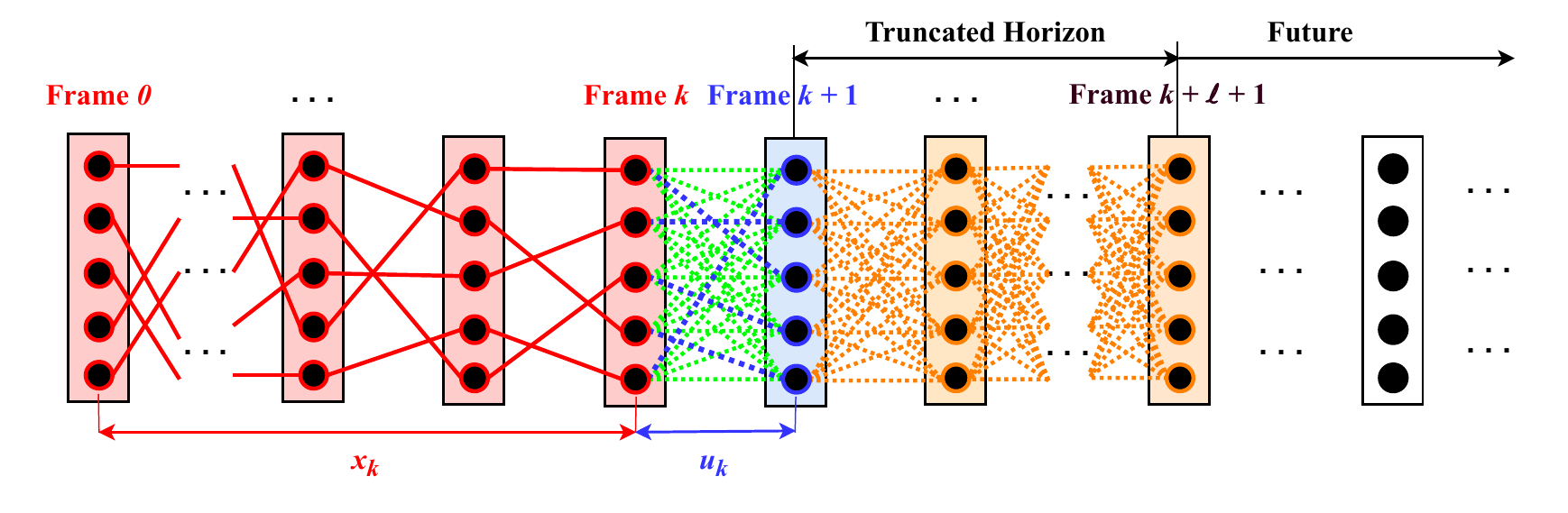}
    \caption{}
    \label{fig:copy-yuchao}
\end{subfigure}
\begin{subfigure}[b]{\textwidth}
    \centering
    \includegraphics[width=0.6\textwidth]{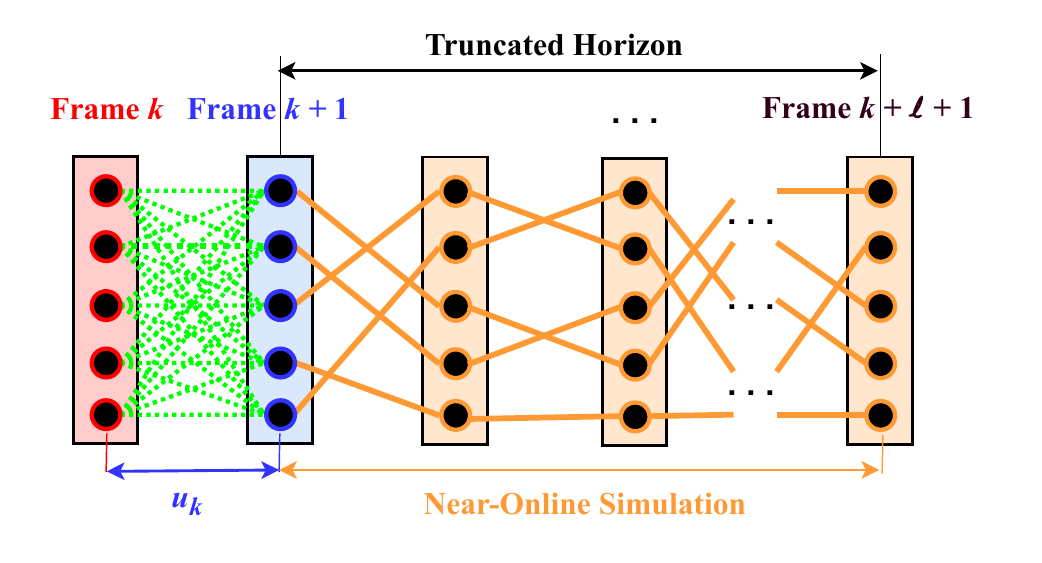}
    \caption{}
    \label{fig:nearonline}
\end{subfigure}
\begin{subfigure}[b]{\textwidth}
    \centering
    \includegraphics[width=\textwidth]{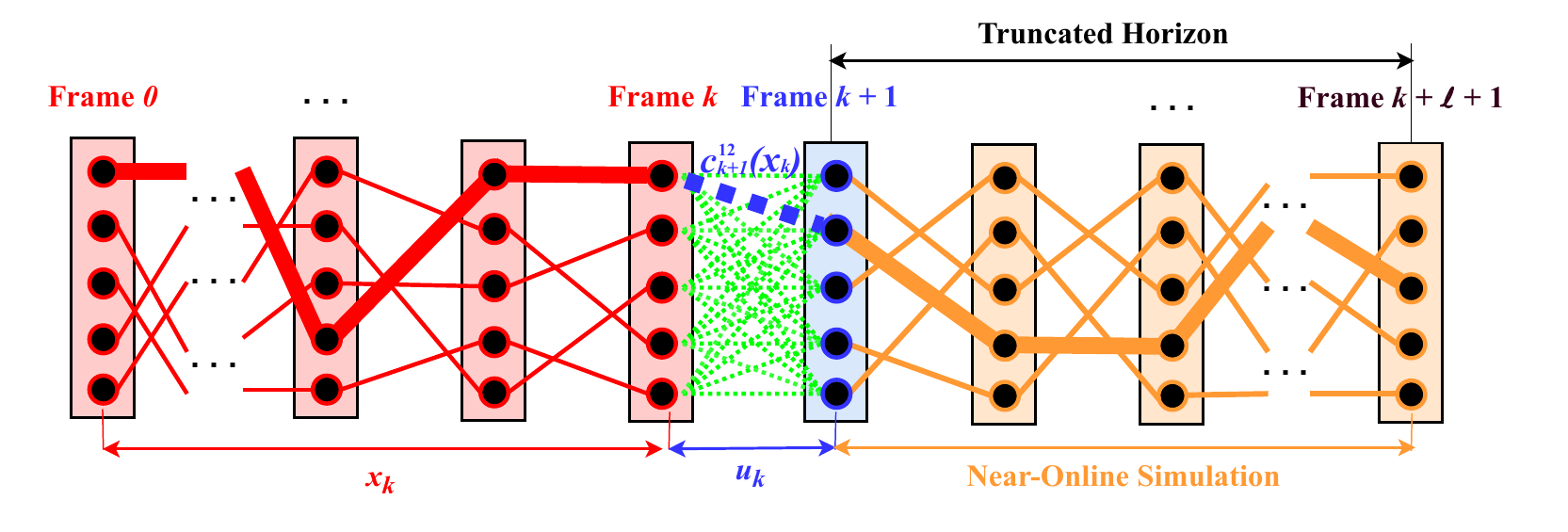}
    \caption{}
    \label{fig:costfuncapprox}
\end{subfigure}
\caption{(a) Overview of key components involved in ADPTrack for MDA (MOT). Red layers and respective arcs represent objects formed in partial groupings (given tracks) up to frame $k$. Blue layer $k+1$ represents the target frame. Green arcs between $k$ and $k+1$ layers represent all pairwise arcs between $\mathcal{N}_k$ and $\mathcal{N}_{k+1}$. Dashed blue matching indicates a selected control $u_k$. Orange layers and arcs represent subsequent frames. 
(b) Visualization of near-online simulation in ADPTrack. The base heuristic is applied to simulate the solution of the MOT problem starting at frame $k+1$ and ending at frame $k+\ell+1$. The obtained tentative tracks are shown by the orange solid lines. 
(c) Illustration of computation of similarity scores $c_{k+1}^{ij}(x_k)$. For example, the weight $c_{k+1}^{12}(x_k)$ is assigned to the bold blue dashed arc, which is dependent on the given track (bold red) and the tentative track (bold orange) that the arc connects.} 
\label{fig:alg}
\end{figure}

The computation of similarity scores relies on tentative tracks, which are obtained {by} solving an MOT problem starting from the $(k+1)$th frame and ending at $(k+\ell+1)$th frame. We refer to the process of solving this problem as the near-online simulation; see Fig.~\ref{fig:nearonline}. 
For objects contained in the $(k+1)$th frame to $(k+\ell+1)$th frame, we treat the $(k+1)$th frame as if it is the first frame, and set $T^j(\Bar{x}_0)=I_{k+1}^j$, where $\Bar{x}^{k+1}_0$ is an artificial initial state starting at $(k+1)$th frame, $j$ is the index associated with objects in the $(k+1)$th frame. We apply the online tracking method to solve the MOT problem starting from the $(k+1)$th frame and ending with $(k+\ell+1)$th frame. The information associated with the tentative track that starts from the $j$th object in $(k+1)$th frame is denoted by $T^j(\Bar{x}^{k+1}_\ell)$, where $\Bar{x}^{k+1}_\ell=(\Bar{u}_{k+1},\dots,\Bar{u}_{k+\ell+1})$. The information contained in $T^j(\Bar{x}^{k+1}_\ell)$ includes the information $I_{k+1}^j$, and will be used to compute the weights associated with the object $j$, as we discuss next. {This is the key to overcome the challenges associated with occlusion. Intuitively, if some object is partially occluded in the target frame, matching based on its visual information in the target frame may lead to error. In contrast, ADPTrack corrects this error by incorporating information in the subsequent frames, where the object may no longer be occluded.} {Moreover, ADPTrack provides a flexible platform for constructing tentative tracks by using arbitrary existing online tracking methods and through online simulation. This is an essential idea in many approximate DP algorithms and sets apart ADPTrack {from} existing near-online methods that often require additional off-line training.}

Suppose that $m$ tracks have been formed for objects contained in $0$th to $k$th frames according to the matching $x_k$. The information associated with each track $i$ is denoted by $T^i(x_k)$. For the tentative tracks formed according to $\Bar{x}^{k+1}_\ell$ via the base heuristic starting from the $j$th object in $(k+1)$th frame, the information associated with it is $T^j(\Bar{x}^{k+1}_\ell)$. The weight term $c_{k+1}^{ij}(x_k)$ is given by
\begin{equation}
    \label{eq:c_def}
    c^{ij}_{k+1}=(1-\alpha) w^{ij}_{k+1}(x_k)+\alpha z^{ij}_{k+1}(x_k),
\end{equation}
where $\alpha$ is a tuning parameter, and $w^{ij}_{k+1}(x_k)$ is given by the base heuristic via \eqref{eq:base_w}, as a similarity measure between objects in $i$th track and $j$th object in the target frame. The term $z^{ij}_{k+1}(x_k)$ reflects the similarity between objects with the information $T^i(x_{k})$ and $T^j(\Bar{x}^{k+1}_\ell)$, and therefore is a similarity measure between given and tentative tracks; see Fig.~\ref{fig:costfuncapprox}. In particular, it is defined as
\begin{equation}
\label{eq:z_def}
z^{ij}_{k+1}(x_k)=F\big(T^i(x_{k}),T^j(\Bar{x}^{k+1}_\ell)\big),
\end{equation}
where {$F$ is a user-defined function, which may depend on the base heuristic. It is used to describe the `similarity' of objects contained in the $i$th given track and the tentative track that starts from object $j$ in $(k+1)$th frame. Intuitively, this helps to overcome the difficulty due to occlusion, as discussed earlier, and a particular implementation is given in Section~\ref{subsec:adptrack_sot}.} 
Note that $\Bar{x}^{k+1}_\ell$ is independent of $x_k$. Once all track-object pairs are computed, we compute the cost function approximation $\Tilde{J}_{k+1}$ according to \eqref{eq:kplus1tilde}, and perform the maximization calculation accordingly. A pseudocode description of ADPTrack for matching one frame is given as Algorithm~\ref{alg:adptrack}.

\begin{algorithm}
\caption{ADPTrack with a Base Heuristic}
\label{alg:adptrack}
    \begin{algorithmic}[1]
    \Statex{\textbf{Input:} Tracks $T^i(x_k),i=1,\dots,m$; target frame $\mathcal{N}_{k+1}$, subsequent frames $\mathcal{N}_{k+2},\dots,\mathcal{N}_{k+\ell+1}$.}
    \Statex{\textbf{Output:} Tracks $T^i(x_{k+1}),\,i=1,2,\ldots,m$.}
    
        \State{Set $w^{ij}_{k+1}(x_{k})\gets H\big(T^i(x_k),I_{k+1}^j\big)$ according to \eqref{eq:base_w} for all $i=1,\dots,m$ and $j\in \mathcal{N}_{k+1}$.}

        \State{Compute $T^j(\bar{x}_\ell^{k+1})$ via simulation the base heuristic on $\mathcal{N}_{k+1},\dots,\mathcal{N}_{k+\ell+1}$, for all $j\in \mathcal{N}_{k+1}$.}
        \State{Set $z^{ij}_{k+1}(x_k)\gets F\big(T^i(x_{k}),T^j(\Bar{x}^{k+1}_\ell)\big)$, for all $i=1,\dots,m$ and $j\in \mathcal{N}_{k+1}$.}
 
        \State{Set $c^{ij}_{k+1}\gets\alpha w^{ij}_{k+1}(x_k)+(1-\alpha)z^{ij}_{k+1}(x_k)$, for all $(i,j)$.}
 
        \State{Compute maximum weight bipartite matching $\tilde{u}_{k}$ with weights given by $c_{k+1}^{ij}(x_k)$; cf. \eqref{eq:approxbellman}.}
      
        \State{Set $x_{k+1}\gets (x_k,\Tilde{u}_k)$, compute $T^i(x_{k+1})$ for $i=1,\dots,m$.}
        
    \end{algorithmic}
\end{algorithm}

\subsection{ADPTrack with BoT-SORT as Base Heuristic}\label{subsec:adptrack_sot}

As an example implementation, we describe ADPTrack with BoT-SORT~\cite{botsort} as a base heuristic. The testing results of this implementation are reported in Section~\ref{sec:experimentalstudy}.
BoT-SORT is an online tracking method that uses a Kalman filter~\cite{kalman} as a motion model and a reidentification neural network~\cite{fastreid} as an appearance model. We apply BoT-SORT directly to generate the tentative tracks in the near-online simulation. Once we obtain the tentative tracks, we compute $z_{k+1}^{ij}(x_k)$ according to \eqref{eq:z_def}. For our numerical studies, $F$ performs an object-to-object visual comparison using the appearance model used in BoT-SORT and computes an average of all the pairwise visual similarity scores to obtain an overall score. In particular, cosine-similarity $\text{(CS)}$ is used in BoT-SORT to extract a visual similarity score between objects, where $\text{CS}$ takes as inputs a visual information pair $(v^i_a,\bar{v}^j_b)$ and returns a high value if they appear `alike' and low otherwise. Roughly speaking, we define the function $F$ as 
\begin{equation}\label{eq:prev_future_score}
   F\big(T^i(x_{k}), T^j(\Bar{x}_{\ell}^{k+1})\big)=\frac{1}{s\cdot \ell}\sum_{v^i_a\in \Bar{V}^i(x_k)}\sum_{\bar{v}^j_{b}\in V^j(\Bar{x}_{\ell}^{k+1})}\textsc{CS}(v^i_a,\bar{v}^j_b),
\end{equation}
where $\Bar{V}^i(x_k)=(v_{q-s+1}^i,v_{q-s+2}^i,\dots,v_q^i)$ represents the high quality visual information of $i$th given track where $q\leq k$, $s$ is a tuning parameter, and $V^j(\Bar{x}_{\ell}^{k+1})=(\Bar{v}^j_{k+1},\Bar{v}^j_{k+2},\dots,\Bar{v}^j_{k+\ell+1})$ represents the entire visual information associated with the tentative track starting from the object $j$ in $\mathcal{N}_{k+1}$. Further details on $F$, the calculation of $q$, and the choice of $s$ are provided in the appendix.

\section{Experimental Study}\label{sec:experimentalstudy}
In this section, we describe the results of our benchmark evaluation that compares ADPTrack using BoT-SORT as a base heuristic, with BoT-SORT itself.
We apply visual modules from BoT-SORT without modification for our experiments. In particular, we use a Yolox model~\cite{yolox} trained by~\cite{bytetrack} for object detection and FastReid's SBS-50 model~\cite{fastreid} fine-tuned by~\cite{botsort} as an appearance model, as has been done for BoT-SORT. 
We ran all our experiments on a V100 20-core GPU.


\textbf{Dataset and Metrics:} We used the MOT17~\cite{milan2016mot16} dataset under the private-detection protocol for evaluating our algorithms. 
We perform our experiments on the second half of the training dataset of the MOT17 dataset, which we refer to as the validation dataset.
This is because the first half of the training dataset has been used by the base heuristic to train the re-identification network~\cite{botsort} (i.e. the appearance model).
Note that our solution does not require any additional training.
We perform our benchmark evaluation on the testing set of the MOT17 dataset.
The videos in the MOT17 dataset consist of many instances where people are temporarily occluded and reappear afterward.

We adopt \emph{clear metrics}~\cite{clear} (among others) to evaluate ADPTrack with respect to the baseline and other state-of-the-art methods. 
More specifically, we describe our results with respect to the following metrics: IDF1~\cite{idf1}, \emph{multi-object tracking accuracy} (MOTA)~\cite{clear}, \emph{higher-order tracking accuracy} (HOTA)~\cite{hota}, \emph{ID-switches} (IDSW), \emph{fragmentations} (Frag), \emph{false positives} (FP), and \emph{false negatives} (FN).
The IDF1 score mainly assesses association accuracy, and MOTA mainly assesses detection accuracy.
IDSW measures the number of incorrect ID switches and Frag measures the number of times a track is missing detections in its trajectory. 
We use Trackeval~\cite{trackeval} to generate the scores according to clear metrics. 
As the focus of ADPTrack is handling temporarily occluded objects, we hypothesize a greater improvement in IDF1 and IDSW metrics.


\textbf{Results:} Table~\ref{tab:comparison-val} shows the overall scores of both tracking methods over the validation dataset, and Table~\ref{tab:comparison-test} shows the overall scores achieved by both of tracking methods over the benchmark dataset.
We present the video-wise improvement scores (i.e., the scores obtained for specific videos) of ADPTrack as compared to the BoT-SORT algorithm for videos in the MOT17 dataset in Table \ref{tab:imp}.
In all tables, the arrow $\uparrow (\downarrow)$ in the first row indicates an increase (resp. decrease) in the corresponding score is more desirable.
We provide the video-wise scores of all other metrics of ADPTrack relative to BoT-SORT on the MOT17 dataset (including graphical illustrations) in the Appendix.
Moreover, we performed ablation studies over various components of ADPTrack.
We also performed parameter variation studies for the tuning parameter, the number of subsequent frames, etc. 
We present these results in the Appendix.
\begin{table}[htbp]
    \centering
    \caption{Comparison of the proposed algorithm with the base heuristic over the validation dataset.}
    \label{tab:comparison-val}
    \begin{tabular}{|l|c|c|c|c|c|c|c|}
        \hline
        Algorithm & IDF1($\uparrow$) & MOTA($\uparrow$) & HOTA($\uparrow$) & FP($\downarrow$) & FN($\downarrow$) & IDSW($\downarrow$)& Frag($\downarrow$)\\ \hline
        BoT-SORT & 83.277 & 80.718 & 70.607 & 9312 & 21432 & 429 & 675\\
        \textbf{ADPTrack} & \color{red}{\textbf{85.355}} & \color{red}{\textbf{81.011}} & \color{red}{\textbf{71.749}} & \color{red}{\textbf{9252}} & \color{red}{\textbf{21090}} & \color{red}{\textbf{357}} & \color{red}{\textbf{657}} \\
        \hline
    \end{tabular}
\end{table}

\begin{table}[htpb]
    \centering
    \caption{Comparison of the proposed algorithm with the base heuristic over the test dataset.}
    \label{tab:comparison-test}
    \begin{tabular}{|l|c|c|c|c|c|c|c|}
        \hline
        Algorithm & IDF1($\uparrow$) & MOTA($\uparrow$) & HOTA($\uparrow$) & FP($\downarrow$) & FN($\downarrow$) & IDSW($\downarrow$)& Frag($\downarrow$)\\ \hline
        BoT-SORT & 80.2 & 80.5 & 65.0 & 22521 & 86037 & 1212 & 1803 \\ 
        \textbf{ADPTrack} & \color{red}{\textbf{80.9}} & \color{red}{\textbf{80.7}} & \color{red}{\textbf{65.4}} & \color{red}{\textbf{22287}} & \color{red}{\textbf{85446}} & \color{red}{\textbf{1086}} & \color{red}{\textbf{1770}} \\
        \hline
    \end{tabular}
\end{table}

\begin{table}[htpb]
    \centering
    \caption{Video-wise improvement of ADPTrack over BoT-SORT for the videos in the MOT17 dataset.}
    \label{tab:imp}
    \begin{tabular}{|l|c|c|c|c|c|c|c|}
        \hline
        Video Name & IDF1($\uparrow$) & MOTA($\uparrow$) & HOTA($\uparrow$) & FP($\downarrow$) & FN($\downarrow$) & IDSW($\downarrow$)& Frag($\downarrow$)\\ \hline
        \textbf{MOT17-01} & \textbf{5.8} & \textbf{0.399} & \textbf{2.9} & \textbf{-7} & \textbf{-9} & \textbf{-6} & \textbf{-1} \\
        \boxit{4.9in}{0.23in} \textbf{MOT17-02} & \textbf{4.923} & \textbf{0.677} & \textbf{3.191} & \textbf{-13} & \textbf{-49} & \textbf{-5} & \textbf{-1} \\ 
        MOT17-03 & 0 & 	\textbf{0.100} & 0 & \textbf{-41} & \textbf{-92} & \textbf{-1} & 1 \\
        MOT17-04 & \textbf{0.141} & -0.012 & -0.156 & 22	& \textbf{-18} &	\textbf{-1} & 2 \\ 
        \boxit{4.9in}{0.08in} \textbf{MOT17-05} & \textbf{1.087} & \textbf{1.012} & \textbf{1.146} & \textbf{-21} & \textbf{-10} & \textbf{-3} & 1 \\ 
        MOT17-06 & -0.89 & \textbf{0.300} & -0.299 &	43 & \textbf{-63} & \textbf{-9} & 4 \\ 
        {\textbf{MOT17-07}} & {\textbf{6.400}} & {\textbf{0.80}} & {\textbf{3.8}}	& {\textbf{-88}} & \textbf{-36} & {\textbf{-15}} & {\textbf{-16}} \\ 
         \textbf{MOT17-08} & \textbf{1.6} & 0 & \textbf{0.799} & \textbf{-10} & 22 & \textbf{-17} & \textbf{-7} \\ 
        \textbf{MOT17-09} & \textbf{6.736} & \textbf{1.633} & \textbf{4.804} & \textbf{-4} & \textbf{-37} & \textbf{-6} & \textbf{-4} \\ 
        \textbf{MOT17-10} & \textbf{5.433} & \textbf{-0.253} & \textbf{3.538} & 17 & 5 & \textbf{-7} & \textbf{-2} \\ 
        \boxit{4.9in}{0.7in} \textbf{MOT17-11} & \textbf{1.899} & \textbf{-0.132} &	\textbf{0.965} & 0 & 8 & \textbf{-2} & \textbf{-1} \\
        MOT17-12 & -2 & -0.5 &	-0.70 &	29 & 8 & 3 & 4 \\
        MOT17-13 & \textbf{0.220} & \textbf{1.077} & \textbf{0.201} & \textbf{-21}	& \textbf{-13} & 0 & \textbf{-1} \\       
        MOT17-14 & -0.900 &	\textbf{0.10} & -0.5 & \textbf{-4} & \textbf{-27} & 3 & 4 \\
        \hline
    \end{tabular}
\end{table}

Over the validation dataset in Table~\ref{tab:comparison-val}, we see an overall improvement of 2.1\% in the IDF1 metric, 0.4\% in the MOTA metric, 1.1\% in the HOTA metric, and a considerable reduction in the FP, FN, IDSW, and Frag metrics. 
In the benchmark evaluation (on the test dataset) in Table~\ref{tab:comparison-test}, we see an overall improvement of 0.7\% in the IDF1 metric, 0.2\% in the MOTA metric, 0.4\% in the HOTA metrics, and a considerable reduction in FP, FN, IDSW, and Frag metrics. 
Our algorithm also outperforms (or is comparable to) several other state-of-the-art tracking methods (see Appendix).

With respect to specific video examples in Table~\ref{tab:imp}, we would like to emphasize the significant improvement of the IDF1 scores in MOT17-01 (5.8\%), MOT17-02 (4.923\%), MOT17-05 (1.08\%), MOT17-07 (6.4\%), MOT17-08 (1.6\%), MOT17-09 (6.736\%), MOT17-10 (5.433\%), and MOT17-11 (1.9\%) videos, which contain many instances of temporary occlusions. 
In the MOT17-03 and MOT17-04 videos, we maintain the accuracy attained by the base heuristic. 
In MOT17-06, 12, and 14, we see a slight reduction in the IDF1 metric (also reflected in the slight increase in IDSW for MOT17-12 and 14).
This increase in error can be attributed to fast-moving cameras. Therefore, the appearance information of the subsequent frames can vary significantly from those that came previously.
ADPTrack with BoT-SORT as the base heuristic achieves an FPS (frames processed per second) rate of 0.8 compared to BoT-SORT which has 4.5.
This is because ADPTrack involves near-online simulations and pairwise comparisons of objects from given and tentative tracks.

\section{Conclusions}\label{sec:discussion}
We have presented an approximate DP framework for rendering existing MOT solution methods more effective and more robust to object occlusions.
The performance of our method on the validation and the test datasets is eminently promising and points to the potential of approximate DP ideas for addressing complex data association problems.
Moreover, the representation of the base heuristic in our model is very general, suggesting that virtually any online tracking method with motion and appearance models can be adapted to improve performance against occlusion. 

With respect to limitations, our solution is more computationally expensive than the base heuristic and introduces delay.
Consequently, there may be practical disadvantages associated with moving from the online to the near-online setting.
These limitations highlight a tradeoff between introducing delays/increasing time complexity and improving the quality of assignments.  
On the other hand, there is an opportunity for parallelization in computing the similarity scores since the computations necessary for each arc are independent. 
Moreover, there are other potential optimizations that we have not implemented, such as storing the appearance similarity scores instead of recomputing them in some cases.

Regarding future work, we hypothesize that generating tentative tracks from frames $k+2$ or beyond may be useful for reducing error, particularly when the object is occluded in frame $k+1$.
This is because an occluded object may negatively influence similarity scores.
Lastly, an interesting avenue for future research is the idea of using ADPTrack with a specified base heuristic as a base heuristic itself.

\printbibliography{}

\appendix
\section{Appendix}
\subsection{Supplemental Results}
\label{appendix:results}
\begin{figure}[H]
    \centering
    \includegraphics[width=\textwidth]{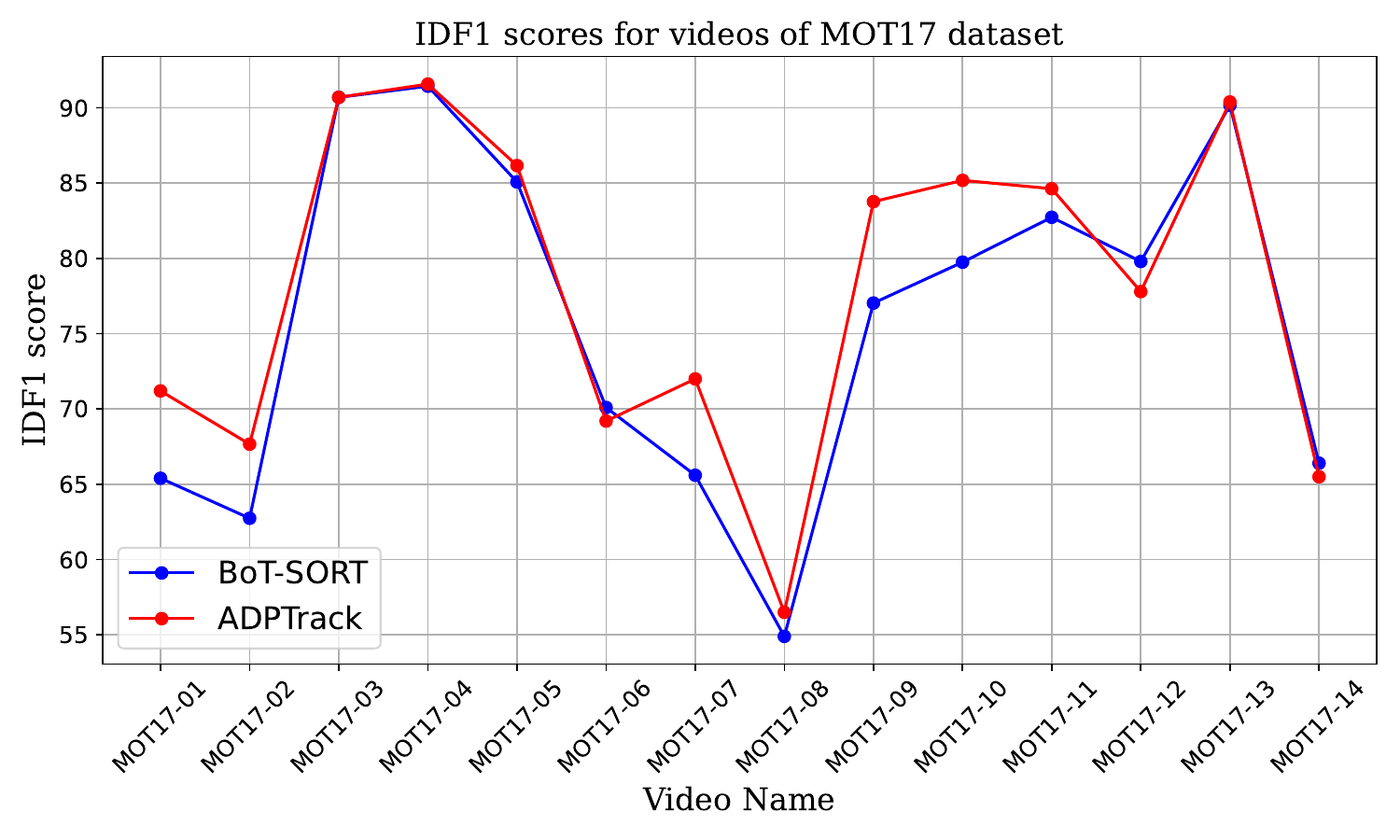}
    \caption{Video-wise IDF1($\uparrow$) scores of BoT-SORT and ADPTrack with BoT-SORT as the base heuristic when applied to videos of the MOT17 dataset.}
    \label{fig:idf1scores}
\end{figure}

\begin{figure}[H]
    \centering
    \includegraphics[width=\textwidth]{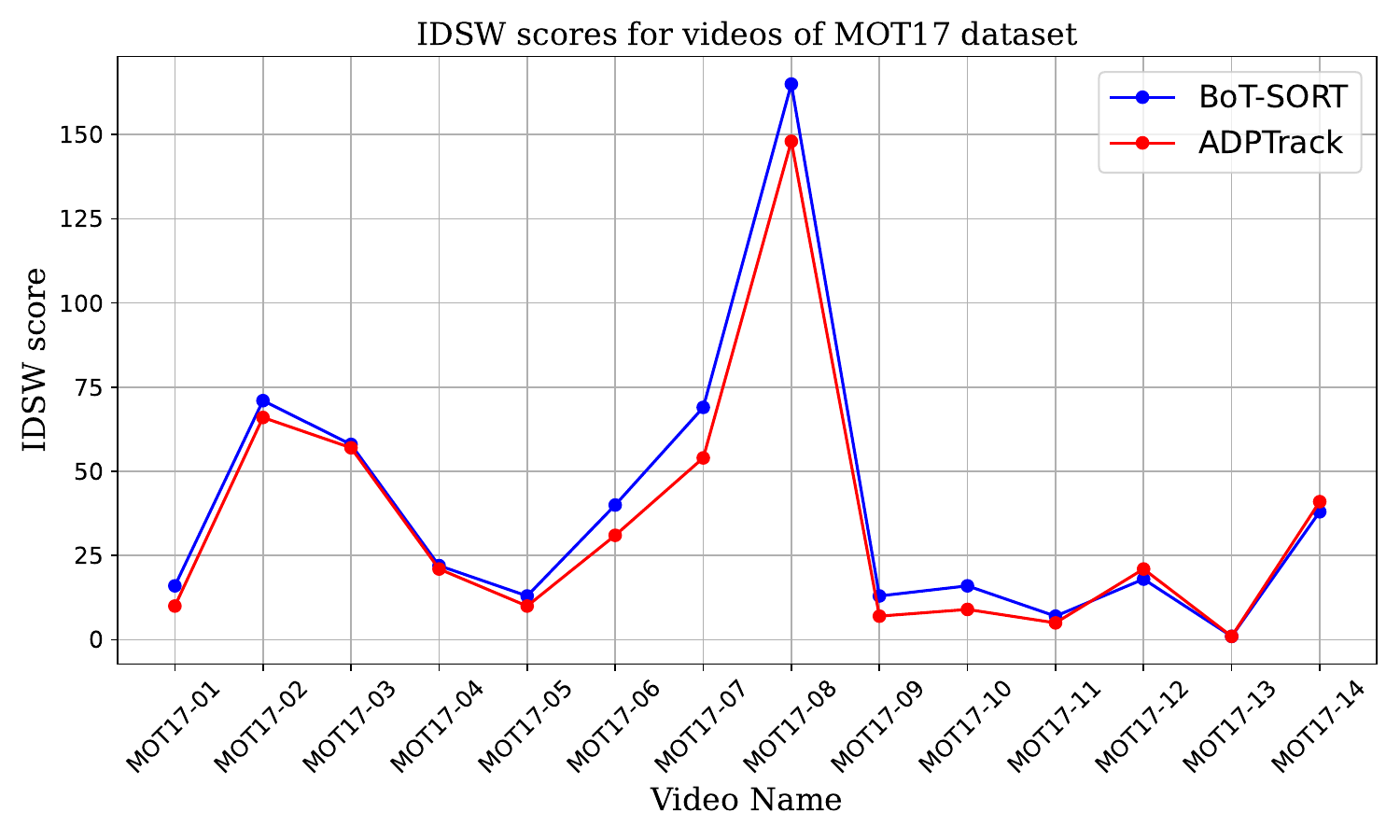}
    \caption{Video-wise IDSW($\downarrow$) scores of BoT-SORT and ADPTrack with BoT-SORT as the base heuristic when applied to videos of the MOT17 dataset.}
    \label{fig:idswscores}
\end{figure}

\begin{figure}[H]
    \centering
    \includegraphics[width=\textwidth]{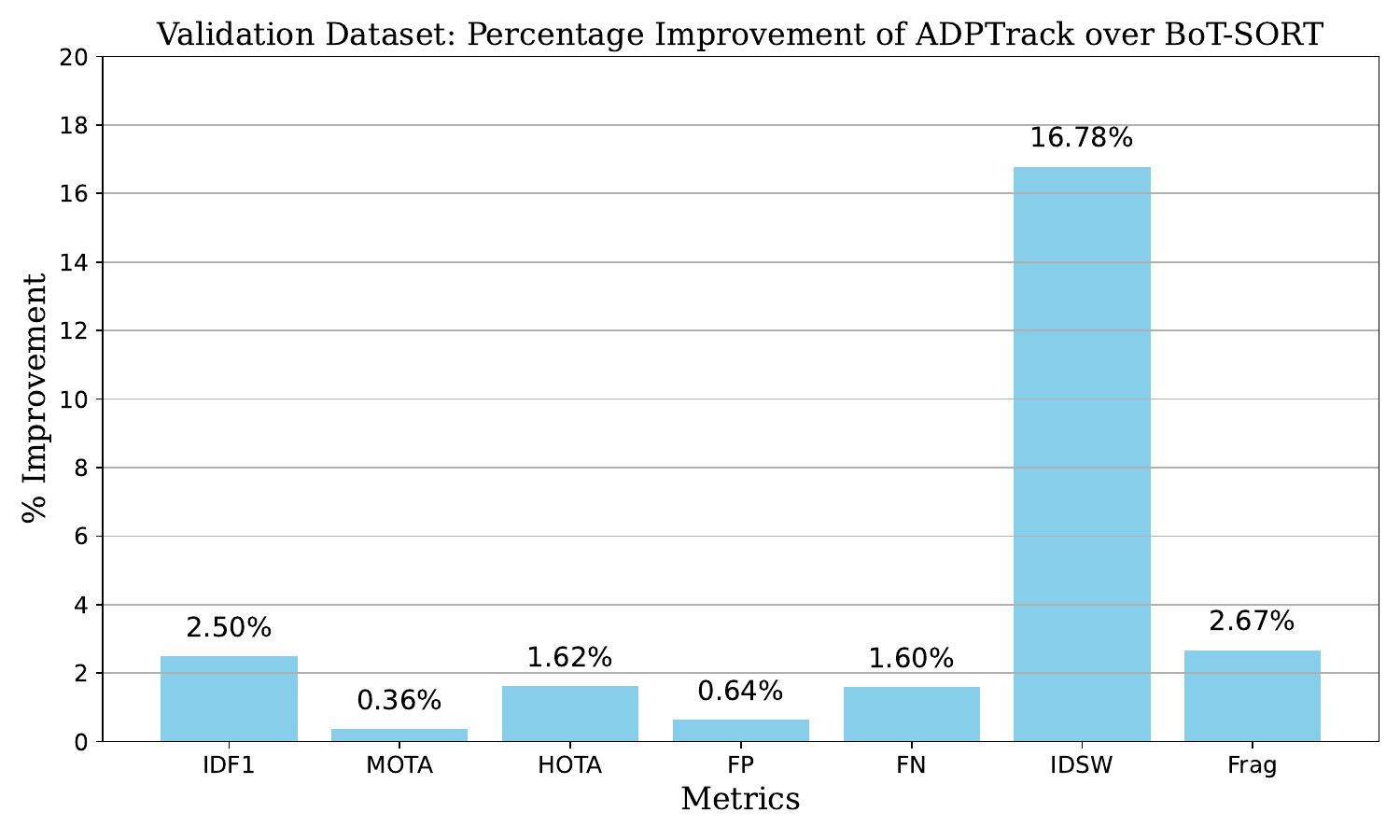}
    \caption{Percentage improvement of ADPTrack with BoT-SORT as the base heuristic over BoT-SORT itself across several MOT metrics on the validation dataset.}
    \label{fig:valimpperc}
\end{figure}

\begin{figure}[H]
    \centering
    \includegraphics[width=\textwidth]{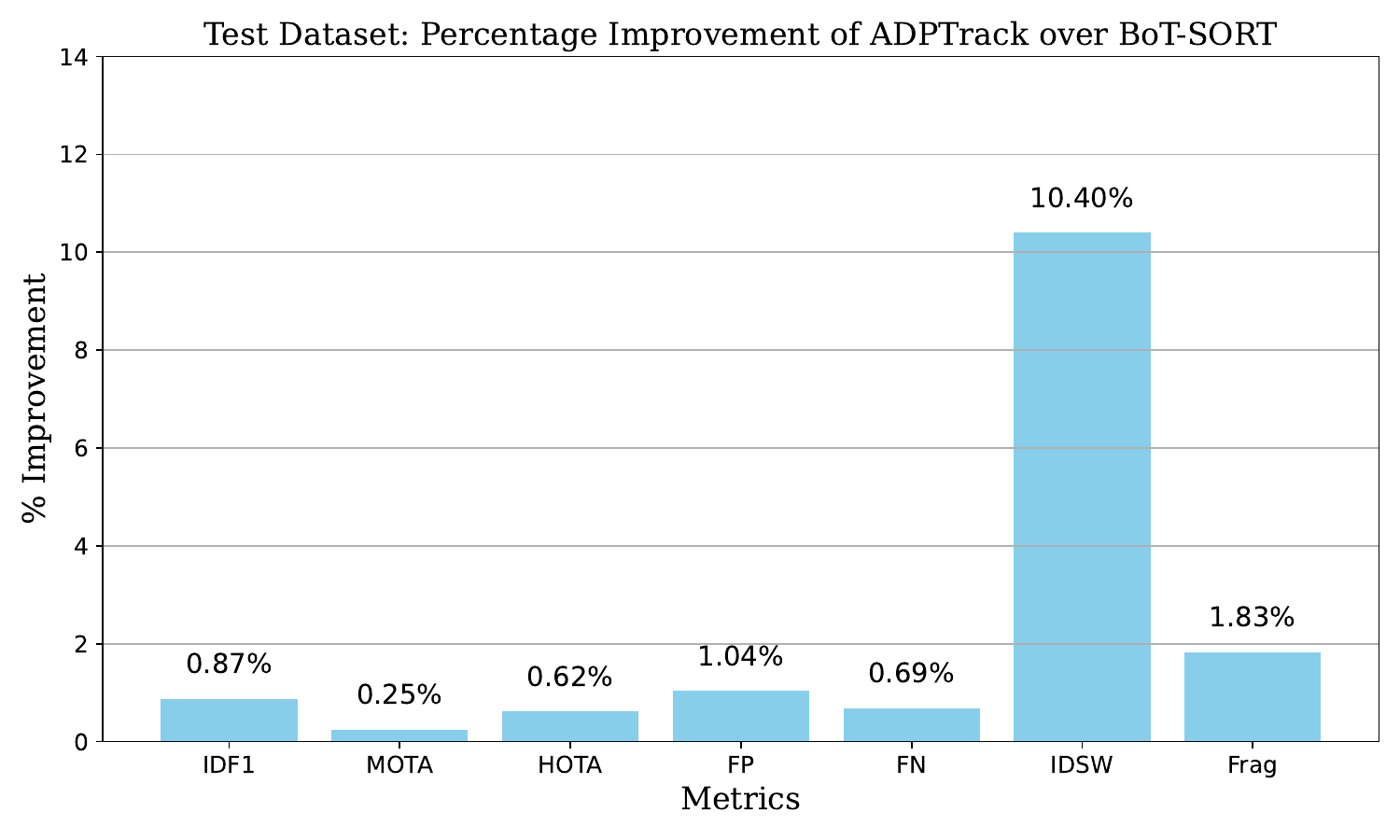}
    \caption{Percentage improvement of ADPTrack with BoT-SORT as the base heuristic over BoT-SORT itself across several MOT metrics on the test dataset.}
    \label{fig:testimpprec}
\end{figure}

We illustrate the IDF1 and IDSW scores of BoT-SORT and ADPTrack with BoT-SORT as the base heuristic for all the videos of the MOT17 dataset in Figs.~\ref{fig:idf1scores} and ~\ref{fig:idswscores}, respectively.
Figs.~\ref{fig:valimpperc} and~\ref{fig:testimpprec} illustrate the percentage improvement of ADPTrack over the base heuristic BoT-SORT across all the metrics over the validation and test dataset, respectively.
We present the video-wise scores of base heuristic and ADPTrack over the validation and test datasets in Tables~\ref{tab:videowise-botsort} and \ref{tab:videowise-adp}, respectively. 
In Table~\ref{tab:other-benchmark}, we compare the overall scores of some state-of-the-art tracking methods with ADPTrack using BoT-SORT as the base heuristic. 
Upon acceptance, we will provide a link to the official scores on the MOT17 Challenge Website.

\begin{table}[H]
    \centering
    \caption{Video-wise scores of BoT-SORT when applied to the MOT17 dataset.}
    \label{tab:videowise-botsort}
    \begin{tabular}{|l|c|c|c|c|c|c|c|}
        \hline
        Video Name & IDF1($\uparrow$) & MOTA($\uparrow$) & HOTA($\uparrow$) & FP($\downarrow$) & FN($\downarrow$) & IDSW($\downarrow$) & Frag($\downarrow$) \\
        \hline
        MOT17-01 & 65.4 & 63.4 & 54.4 & 461 & 1883 & 16 & 31 \\ 
        MOT17-02 & 62.743 & 60.152 & 52.711 & 1007 & 2859 & 71 & 92 \\
        MOT17-03 & 90.7 & 92.6 & 73.1 & 3942 & 3789 & 58 & 119\\
        MOT17-04 & 91.44 & 91.008 & 79.886 & 894 & 1258 & 22 & 31 \\
        MOT17-05 & 85.08 & 82.306 & 67.458 & 221 & 360 & 13 & 14 \\
        MOT17-06 & 70.1 & 66.6	& 56.9 & 959 & 2933 & 40 & 73 \\ 
        MOT17-07 & 65.6 & 74.6 & 53.9 & 550 & 3674 & 69 & 106 \\ 
        MOT17-08 & 54.9 & 65.9 & 48.5 & 1059 & 5977 & 165 & 184 \\ 
        MOT17-09 & 77.036 & 86.662 & 65.876 & 26 & 345 & 13 & 11 \\
        MOT17-10 & 79.753 & 74.911 & 59.841 & 240 & 1230 & 16 & 57 \\
        MOT17-11 & 82.733 & 71.773 & 70.984 & 611 & 657 & 7 & 14 \\
        MOT17-12 & 79.8 & 72.2 & 64.2 & 291	& 2104 & 18 & 34 \\         
        MOT17-13 & 90.171 & 82.858 & 71.038 & 105 & 435 & 1 & 6 \\
        MOT17-14 & 66.4 & 53.5 & 48.3 & 245	& 8319 & 38 & 54 \\ 
        \hline
    \end{tabular}
\end{table}
\begin{table}[H]
    \centering
    \caption{Video-wise scores of ADPTrack with BoT-SORT as a base heuristic when applied to the MOT17 dataset.}
    \label{tab:videowise-adp}
    \begin{tabular}{|l|c|c|c|c|c|c|c|}
        \hline
        Video Name & IDF1($\uparrow$) & MOTA($\uparrow$) & HOTA($\uparrow$) & FP($\downarrow$) & FN($\downarrow$) & IDSW($\downarrow$) & Frag($\downarrow$) \\ \hline
        MOT17-01 & 71.2 & 63.8 & 57.3 & 454 & 1874 & 10 & 30 \\
        MOT17-02 & 67.666 & 60.83 & 55.902 & 994 & 2810 & 66 & 91 \\ 
        MOT17-03 & 90.7 & 92.7 & 73.1 & 3901 & 3697 & 57 & 120\\ 
        MOT17-04 & 91.581 & 90.996 & 79.729 & 916 & 1240 & 21 & 33 \\ 
        MOT17-05 & 86.167 & 83.318 & 68.604 & 200 & 350 & 10 & 15 \\ 
        MOT17-06 & 69.2 & 66.9 & 56.6 & 1002 & 2870 & 31 & 77 \\ 
        MOT17-07 & 72.0 & 75.4 & 57.7 & 462 & 3638 & 54 & 90\\ 
        MOT17-08 & 56.5 & 65.9 & 49.3 & 1049 &	5999 & 148 & 177 \\
        MOT17-09 & 83.772 & 88.295 & 70.681 & 22 & 308 & 7 & 7 \\ 
        MOT17-10 & 85.186 & 74.658 & 63.379 & 257 & 1235 & 9 & 55 \\ 
        MOT17-11 & 84.633 & 71.64 & 71.949 & 611 & 665 & 5 & 13 \\ 
        MOT17-12 & 77.8 & 71.7 & 63.5 & 320	& 2112 & 21 & 38 \\ 
        MOT17-13 & 90.392 & 83.935 & 71.24 & 84 & 422 & 1 & 5 \\ 
        MOT17-14 & 65.5 & 53.6 & 47.8 & 241 & 8292 & 41 & 58 \\        
        \hline
    \end{tabular}
\end{table}
\begin{table}[H]
    \centering
    \caption{Overall scores of ADPTrack with BoT-SORT as a base heuristic, BoT-SORT itself, and other state-of-the-art tracking methods on the MOT17 test dataset.}
    \label{tab:other-benchmark}
    \begin{tabular}{|l|c|c|c|c|c|c|c|}
        \hline
        Method & IDF1($\uparrow$) & MOTA($\uparrow$) & HOTA($\uparrow$) & FP($\downarrow$) & FN($\downarrow$) & IDSW($\downarrow$) & Frag($\downarrow$) \\ \hline
        OCSORT~\cite{ocsort} & 77.5 & 78.0 & 63.2 & \textbf{15129} & 107055 & 1950 & 2040 \\
        Deep-OCSORT~\cite{deepocsort} & 80.6 & 79.4 & 64.9 & 16572 & 98796 & \textbf{1023} & 2196 \\ 
        StrongSORT++~\cite{strongsort} & 79.5 & 79.6 & 64.4 & 27876 & 86205 & 1194 & 1866 \\
        ByteTrack~\cite{bytetrack} & 77.3 & 80.3 & 63.1 & 25491 & 83721 & 2196 & 2277 \\
        MotionTrack~\cite{motiontrack} & 80.1 & \textbf{81.1} & 65.1 & 23802 & 81660 & 1140 & \textbf{1605} \\
        UTM~\cite{utm} & 78.7 & 81.8 & 64.0 & 25077 & \textbf{76298} & 1431 & 1889 \\
        BoT-SORT~\cite{botsort} & 80.2 & 80.5	& 65.0 & 22521 & 86037 & 1212 & 1803 \\ 
        ADPTrack & \textbf{80.9} & 80.7 & \textbf{65.4} & 22287 & 85446 & 1086 & 1770 \\ 
        \hline
    \end{tabular}
\end{table}
\subsection{Ablation Studies}
\label{appendix:ablation}
This subsection describes several ablation studies performed for all the components we introduced in the ADPTrack framework.
We list the experimental studies for several components and compare their results to the chosen setup for our algorithm presented in Section~\ref{sec:algorithm}. 
In the first experiment, we use a simplified base heuristic with only the motion model and no appearance model.
In this experiment, we test the importance of the appearance model in our framework.
In the second experiment, we consider the standalone base heuristic and compare the visual information of the target frame objects with all the matched objects of the given tracks without our near-online simulation approach. 
This experiment tests the importance of using tentative tracks in our framework.
In the third experiment, we use ADPTrack with a base heuristic that has both the motion and appearance models, but we use a modified similarity score [cf. \eqref{eq:z_def}].
In this experiment, we test the importance of using visual information of previously matched objects directly instead of using overall averaged visual information for a given track $i$ while comparing it with a tentative track $j$.
In the fourth experiment, we present additional implementation details related to the main algorithm~\ref{alg:adptrack}.
Every experiment includes a parameter sweep.

\subsubsection{Experiment 1}
The first experiment tests as the base heuristic the BoT-SORT method without its appearance model. Since it relies solely on the motion model, we refer to it as the BoT-SORT-motion. 
In particular, we apply the concept of intersection-over-union overlap (IOU overlap for short) used by BoT-SORT to compute similarity scores.
While performing assignments between the given tracks and the objects of the target frame, BoT-SORT uses a motion model to predict the position and size of the object of a given track in the next frame. 
It then computes an IOU overlap between the predicted position and size of the object and the objects of the target frame based on the position and size of the objects. 
The IOU overlap scores are used to perform the assignment between the given tracks and the target frame.

In this ablation study, each of the tentative track's objects is sequentially compared to the given track's object in the subsequent frames using IOU overlap to obtain a sequence of IOU overlap scores.
We calculate $z_{k+1}^{ij}(x_k)$ by taking an average of $\ell$ IOU overlap scores associated with the $\ell$ objects in tentative track $j$ obtained previously.
Note that we maintain $m$ copies of a given track's motion model to generate $m$ different $z_{k+1}^{ij}(x_k)$ similarity scores for a single track $i$.

More specifically, we use a motion model, which predicts the position and size of the object based on historical information. Given the position and size information $(p_0^i,p_1^i,\dots,p_k^i)$ associated with the $i$th track, we predict the position and size information $\Tilde{p}_{k+1}^i$, 
\begin{equation}
    \label{eq:track_p}
    \Tilde{p}_{b+1}^i=\text{MOTION}(p_0^i,p_1^i,\dots,p_b^i),\quad b=1,\dots,k.
\end{equation}
In our case, $\text{MOTION}$ represents the calculation related to the Kalman filter. Therefore, the information $\Tilde{p}_{k+1}^i$ can be used to represent the size and position of the object at $k+1$ the frame based on the information associated with track $i$. To compute the similarity score of given track $i$ and tentative track starting from $j$ without appearance information, we then compute a sequence of information $\Tilde{p}^{ij}_{k},\Tilde{p}^{ij}_{k+1},\dots,\Tilde{p}^{ij}_{k+\ell}$ via
\begin{equation}
    \label{eq:tentative_p}
    \Tilde{p}_{b+1}^{ij}=\text{MOTION}(p_0^i,p_1^i,\dots,p_k^i,\Bar{p}_{k+1}^{j},\dots,\Bar{p}_b^{j}),\quad b=k+1,\dots,k+\ell+1,
\end{equation}
In addtion, we define $\Tilde{p}_{k+1}^{ij}=\Tilde{p}_{k+1}^{i}$. Based on the obtained sequence $\Tilde{p}^{ij}_{k+1},\dots,\Tilde{p}^{ij}_{k+\ell},\Tilde{p}^{ij}_{k+\ell+1}$, we compute the similarity score for this ablation study and refer to it as $F_1$. 
It is given by
\begin{equation}\label{eq:f1_score}
   F_1\big(T^i(x_{k}), T^j(\Bar{x}_{\ell}^{k+1})\big)=\frac{1}{ \ell}\sum_{b=k+1}^{k+\ell+1}\text{IOU}(\tilde{p}_{b}^{ij},\bar{p}^j_b),
\end{equation}
where $(\Bar{p}^j_{k+1},\Bar{p}^j_{k+2},\dots,\Bar{p}^j_{k+\ell+1})$ represents all the position information associated with the tentative track starting from the object $j$. 
The function $\text{IOU}$ takes two position and size information $\tilde{p}_b^{ij},\bar{p}^j_b$ and returns a big value if `the overlap' of objects represented by $\tilde{p}_b^{ij},\bar{p}^j_b$ is large.

As the base heuristic for this experiment is BoT-SORT-motion, we apply BoT-SORT-motion as a baseline for comparison. Table \ref{tab:baseline} and \ref{tab:experiment1} show the video-wise results of BoT-SORT-motion and ADPTrack with BoT-SORT-motion as the base heuristic for all the videos of the validation dataset.  
The overall scores for BoT-SORT-motion and ADPTrack with BoT-SORT-motion as the base heuristic are mentioned in Table \ref{tab:comparison-val-ablationstudies-exp1}. 

We see a significant improvement in the IDF1 metric for MOT17-02 (4.04\%), MOT17-04 (0.78\%), MOT17-05 (1.948\%), and MOT17-11 (4.093\%). 
In fact, out of all the experiments, we see the best score for the MOT17-04 video in this experiment, which is 92.236\%. 
However, we see a reduction in IDF1 accuracies for MOT17-09, MOT17-10, and MOT17-13 which may be because of a moving camera. 
We believe that this kind of tracker is suitable for videos in which the motion of the object can be more helpful, such as MOT17-04 with an overhead static camera. 
Across all the videos, we see an overall improvement of 1.35\% in the IDF1 metric, 0.5\% in the HOTA metric, and better IDSW, FP, FN, and Frag scores at a slightly better MOTA metric.
We believe that this is quite promising, given that it is only using a motion model, which can be quite inexpensive.
\begin{table}[H]
    \centering
    \caption{Video-wise scores of BoT-SORT-motion when applied to the validation dataset.}
    \label{tab:baseline}
    \begin{tabular}{|l|c|c|c|c|c|c|c|}
        \hline
        Video Name & IDF1($\uparrow$) & MOTA($\uparrow$) & HOTA($\uparrow$) & FP($\downarrow$) & FN($\downarrow$) & IDSW($\downarrow$) & Frag($\downarrow$) \\
        \hline
        MOT17-02 & 60.09 & 59.615 & 51.484 & 1060 & 2867 & 63 & 84 \\
        MOT17-04 & 91.448 & 90.893 & 80.002 & 949 & 1236 & 17 & 30 \\
        MOT17-05 & 81.218 & 82.038 & 65.948 & 223 & 367 & 13 & 16 \\
        MOT17-09 & 83.407 & 86.419 & 70.049 & 23 & 357 & 11 & 10 \\
        MOT17-10 & 77.879 & 75.063 & 58.677 & 239 & 1222 & 16 & 55 \\
        MOT17-11 & 80.53 & 71.596 & 69.233 & 614 & 660 & 9 & 14 \\
        MOT17-13 & 90.689 & 83.08 & 71.354 & 90 & 441 & 3 & 6 \\
        \hline
    \end{tabular}
\end{table}

\begin{table}[H]
    \centering
    \caption{Video-wise scores of ADPTrack as per experiment 1 over the validation dataset.}
    \label{tab:experiment1}
    \begin{tabular}{|l|c|c|c|c|c|c|c|}
        \hline
        Video Name & IDF1($\uparrow$) & MOTA($\uparrow$) & HOTA($\uparrow$) & FP($\downarrow$) & FN($\downarrow$) & IDSW($\downarrow$) & Frag($\downarrow$) \\ \hline
        MOT17-02 & 64.132 & 59.919 & 53.049 & 1118 & 2784 & 58 & 86 \\ 
        MOT17-04 & 92.236 & 90.847 & 80.272 & 940 & 1256 & 17 & 31 \\ 
        MOT17-05 & 83.076 & 82.335 & 65.993 & 235 & 343 & 15 & 20 \\ 
        MOT17-09 & 82.522 & 86.732 & 69.962 & 27 & 344 & 11 & 10 \\ 
        MOT17-10 & 77.093 & 75.131 & 58.459 & 240 & 1216 & 17 & 56 \\ 
        MOT17-11 & 84.623 & 71.618 & 71.861 & 612 & 665 & 5 & 13 \\ 
        MOT17-13 & 90.369 & 83.112 & 71.236 & 89 & 441 & 3 & 6 \\ 
        \hline
    \end{tabular}
\end{table}

\begin{table}[H]
    \centering
    \caption{Comparison of overall scores' of BoT-SORT-motion, and ADPTrack with BoT-SORT-motion as the base heuristic as per experiment 1 over the validation dataset.}
    \label{tab:comparison-val-ablationstudies-exp1}
    \begin{tabular}{|l|c|c|c|c|c|c|c|}
        \hline
        Algorithm & IDF1($\uparrow$) & MOTA($\uparrow$) & HOTA($\uparrow$) & FP($\downarrow$) & FN($\downarrow$) & IDSW($\downarrow$) & Frag($\downarrow$) \\ \hline
        BoT-SORT-motion & 82.55 & 80.553 & 70.346 & 9594 & 21450 & 396 & 645 \\
        Experiment 1 & 83.908 & 80.635 & 70.878 & 9783 & 21147 & 378 & 666 \\ 
        \hline
    \end{tabular}
\end{table} 

\subsubsection{Parameter Study for Experiment 1}
The number of subsequent frames $\ell$ for near-online simulation and the tuning parameter $\alpha$ of \eqref{eq:c_def} are varied in two ablation studies and the experimentation results are presented in Tables \ref{tab:experiment1-ablationstudy} and \ref{tab:exp1-ablation2}, respectively. We observe that the performance of ADPTrack increases as we reach a certain number of subsequent frames, and then decreases as we go further. 

\begin{table}[H]
\centering
\caption{An ablation study over the number of subsequent frames; ADPTrack with BoT-SORT-motion as the base heuristic as per experiment 1 over the validation dataset; $\ell$: number of subsequent frames for near-online simulation.}
\label{tab:experiment1-ablationstudy}
\begin{tabular}{|l|c|c|c|c|c|c|c|}
\hline
$\ell$ & IDF1($\uparrow$) & MOTA($\uparrow$) & HOTA($\uparrow$) & FP($\downarrow$) & FN($\downarrow$) & IDSW($\downarrow$) & Frag($\downarrow$) \\
\hline
1 & 83.449 & 80.49 & 70.567 & 9864 & 21285 & 393 & 672  \\
2 & 83.566 & 80.403 & 70.677 & 9879 & 21417 & 387 & 663  \\
3 & 83.627 & 80.494 & 70.775 & 9906 & 21246 & 384 & 666  \\
4 & 83.782 & 80.64 & 70.771 & 9786 & 21132 & 381 & 666  \\
5 & 83.738 & 80.598 & 70.742 & 9786 & 21204 & 378 & 660  \\
6 & 83.878 & 80.59 & 70.851 & 9783 & 21219 & 378 & 663  \\
7 & 83.908 & 80.635 & 70.878 & 9783 & 21147 & 378 & 666  \\
8 & 83.861 & 80.596 & 70.797 & 9738 & 21249 & 384 & 672  \\
9 & 83.861 & 80.594 & 70.796 & 9741 & 21249 & 384 & 672  \\
10 & 83.82 & 80.551 & 70.824 & 9768 & 21291 & 384 & 666  \\
11 & 83.796 & 80.51 & 70.825 & 9765 & 21363 & 381 & 660  \\
12 & 83.781 & 80.501 & 70.82 & 9741 & 21396 & 387 & 666  \\
13 & 83.78 & 80.512 & 70.827 & 9741 & 21378 & 387 & 663  \\
14 & 83.649 & 80.503 & 70.754 & 9750 & 21381 & 390 & 663  \\
15 & 83.657 & 80.572 & 70.748 & 9759 & 21261 & 390 & 669  \\ 
\hline
\end{tabular}
\end{table}

\begin{table}[H]
\centering
\caption{An ablation study over the tuning parameter $\alpha$; ADPTrack with BoT-SORT-motion as a base-heuristic as per experiment 1 over the validation dataset; $\ell$: number of subsequent frames for near-online simulation.}
\label{tab:exp1-ablation2}
\begin{tabular}{|l|l|c|c|c|c|c|c|c|}
\hline
$\ell$ & $\alpha$ & IDF1($\uparrow$) & MOTA($\uparrow$) & HOTA($\uparrow$) & FP($\downarrow$) & FN($\downarrow$) & IDSW($\downarrow$) & Frag($\downarrow$) \\
\hline
6 & 0.05 & 83.214 & 80.549 & 70.671 & 9600 & 21447 & 399 & 648  \\
6 & 0.1 & 83.694 & 80.685 & 70.879 & 9525 & 21330 & 372 & 648  \\
6 & 0.15 & 83.878 & 80.59 & 70.851 & 9783 & 21219 & 378 & 663  \\
6 & 0.2 & 83.361 & 80.215 & 70.511 & 10023 & 21513 & 450 & 702  \\
7 & 0.05 & 83.214 & 80.549 & 70.671 & 9600 & 21447 & 399 & 648  \\
7 & 0.1 & 83.718 & 80.698 & 70.893 & 9525 & 21312 & 369 & 645  \\
7 & 0.15 & 83.908 & 80.635 & 70.878 & 9783 & 21147 & 378 & 666  \\
7 & 0.2 & 83.599 & 80.282 & 70.648 & 9918 & 21528 & 432 & 699  \\
8 & 0.05 & 83.215 & 80.551 & 70.67 & 9597 & 21447 & 399 & 648  \\
8 & 0.1 & 83.718 & 80.698 & 70.893 & 9525 & 21312 & 369 & 645  \\
8 & 0.15 & 83.861 & 80.596 & 70.797 & 9738 & 21249 & 384 & 672  \\
8 & 0.2 & 83.513 & 80.154 & 70.562 & 10080 & 21546 & 459 & 702  \\
9 & 0.05 & 83.215 & 80.551 & 70.671 & 9597 & 21447 & 399 & 648  \\
9 & 0.1 & 83.726 & 80.679 & 70.928 & 9534 & 21330 & 372 & 648  \\
9 & 0.15 & 83.861 & 80.594 & 70.796 & 9741 & 21249 & 384 & 672  \\
9 & 0.2 & 83.094 & 80.109 & 70.531 & 10161 & 21552 & 444 & 705  \\
10 & 0.05 & 83.215 & 80.551 & 70.671 & 9597 & 21447 & 399 & 648  \\
10 & 0.1 & 83.74 & 80.649 & 70.933 & 9531 & 21378 & 375 & 648  \\
10 & 0.15 & 83.82 & 80.551 & 70.824 & 9768 & 21291 & 384 & 666  \\
10 & 0.2 & 82.433 & 80.174 & 70.114 & 9990 & 21615 & 447 & 714  \\
\hline
\end{tabular}
\end{table}

\subsubsection{Experiment 2}\label{appendix:ablationexp2}
In experiment 2, we neither perform the near-online simulation nor generate the tentative tracks.
We consider BoT-SORT with both the motion and appearance models and exploit the visual information of previously matched objects of given tracks. 
For a given track $i$, we maintain an appearance quality vector where each value indicates the visual quality of a previously matched object.
We generate the appearance quality vector by extending it every time an object is assigned to the given track $i$.
Based on the appearance quality vector and a particular threshold referred to as the quality threshold $\beta$, we iterate backward starting at frame $k$ and find the object at frame $q$ with the appearance quality value greater than the quality threshold $\beta$. Then we select the visual information of objects from frame $q-s+1$ to frame $q$ assigned to track $i$, which is denoted by $\Bar{V}^i(x_k)$, i.e.,  $\Bar{V}^i(x_k)=(v_{q-s+1}^i,v_{q-s+2}^i,\dots,v_q^i)$. The high-quality visual information $\Bar{V}^i(x_k)$ is then used to compute the similarity score.

To compute $z_{k+1}^{ij}(x_k)$, we compare the visual information $v_a^i$ contained in $\Bar{V}^i(x_k)$ with $\Bar{v}_{k+1}^j$ to generate a sequence of visual similarity scores between given track $i$ and object $j$.
We then calculate the average of those visual similarity scores.
In particular, we use a modified similarity score $F$ for this ablation study and refer to it as $F_2$. It is given by
\begin{equation}\label{eq:f2_score}
   F_2\big(T^i(x_{k}), T^j(\Bar{x}_{\ell}^{k+1})\big)=\frac{1}{s}\sum_{v^i_a\in \Bar{V}^i(x_k)}\textsc{CS}(v^i_a,\bar{v}^j_{k+1}).
\end{equation}

The number $s$ of previously matched objects from consecutive frames of given tracks to be used for comparison is flexible and the ablation study for this parameter is shown in Table \ref{tab:experiment2-ablationstudy}. 
We set the value of the quality threshold $\beta$ as 0.15 based on empirical results.
The video-wise results for BoT-SORT and the modified BoT-SORT as per experiment 2 are presented in Tables \ref{tab:videowise-botsort} and \ref{tab:experiment2}, respectively.
The overall scores for BoT-SORT and experiment 2 are listed in Table \ref{tab:comparison-val-ablationstudies}. 

\begin{table}[H]
\centering
\caption{An ablation study over the number of previously matched objects ($\#s$) as per experiment 2.}
\label{tab:experiment2-ablationstudy}
\begin{tabular}{|l|c|c|c|c|c|c|c|}
\hline
$\#s$ & IDF1($\uparrow$) & MOTA($\uparrow$) & HOTA($\uparrow$) & FP($\downarrow$) & FN($\downarrow$) & IDSW($\downarrow$) & Frag($\downarrow$) \\
\hline
1 & 84.59 & 80.956 & 71.321 & 9321 & 21099 & 369 & 645  \\
2 & 84.567 & 80.95 & 71.312 & 9324 & 21102 & 372 & 651  \\
3 & 84.43 & 80.87 & 71.213 & 9402 & 21144 & 381 & 660  \\
4 & 84.43 & 80.87 & 71.213 & 9402 & 21144 & 381 & 660  \\
5 & 84.43 & 80.87 & 71.213 & 9402 & 21144 & 381 & 660  \\
6 & 84.43 & 80.87 & 71.213 & 9402 & 21144 & 381 & 660  \\
7 & 84.43 & 80.87 & 71.213 & 9402 & 21144 & 381 & 660  \\
8 & 84.43 & 80.87 & 71.213 & 9402 & 21144 & 381 & 660  \\
9 & 84.436 & 80.891 & 71.219 & 9378 & 21144 & 372 & 660  \\
10 & 84.436 & 80.891 & 71.219 & 9378 & 21144 & 372 & 660  \\
11 & 84.436 & 80.891 & 71.219 & 9378 & 21144 & 372 & 660  \\
12 & 84.436 & 80.891 & 71.219 & 9378 & 21144 & 372 & 660  \\
13 & 84.436 & 80.891 & 71.219 & 9378 & 21144 & 372 & 660  \\
14 & 84.436 & 80.891 & 71.219 & 9378 & 21144 & 372 & 660  \\
15 & 84.436 & 80.891 & 71.219 & 9378 & 21144 & 372 & 660  \\

\hline
\end{tabular}
\end{table}

\begin{table}[H]
    \centering
    \caption{Video-wise scores for modified BoT-SORT (experiment 2) over the validation dataset.}
    \label{tab:experiment2}
    \begin{tabular}{|l|c|c|c|c|c|c|c|}
        \hline
        Video Name & IDF1($\uparrow$) & MOTA($\uparrow$) & HOTA($\uparrow$) & FP($\downarrow$) & FN($\downarrow$) & IDSW($\downarrow$) & Frag($\downarrow$) \\ \hline
        MOT17-02 & 64.803 & 60.992 & 54.481 & 995 & 2793 & 66 & 90 \\ 
        MOT17-04 & 91.586 & 91.054 & 79.746 & 908 & 1235 & 20 & 32 \\ 
        MOT17-05 & 85.015 & 82.306 & 67.338 & 224 & 358 & 12 & 14 \\ 
        MOT17-09 & 81.787 & 87.669 & 68.861 & 26 & 321 & 8 & 6 \\ 
        MOT17-10 & 84.241 & 74.726 & 62.828 & 258 & 1228 & 11 & 55 \\ 
        MOT17-11 & 84.62 & 71.618 & 71.872 & 611 & 666 & 5 & 13 \\ 
        MOT17-13 & 90.528 & 83.587 & 71.289 & 85 & 432 & 1 & 5 \\ 
        \hline
    \end{tabular}
\end{table}

\subsubsection{Experiment 3}

In this experiment, we use BoT-SORT with both the motion and appearance models as the base heuristic. 
BoT-SORT maintains representative visual information associated with every given track. 
For a given track $i$, the representative visual information is updated whenever an object from a target frame is assigned to the given track $i$.
It is updated by averaging the assigned object's visual information with the given track's representative visual information through an exponential moving average. 
While performing assignments between the given tracks and the objects of the target frame with respect to the appearance model, BoT-SORT compares the representative visual information of a given track $i$ to the visual information of every object of the target frame.

In this ablation study, to calculate a similarity score between a given track $i$ and a tentative track $j$, we compare the representative visual information of the given track $i$ and the visual information of objects of tentative track $j$.
As there are $\ell$ objects in tentative track $j$, this comparison generates a sequence of $\ell$ visual similarity scores.
We also calculate IOU overlap scores as explained in experiment 1.
Similar to BoT-SORT, we consider a minimum of the visual similarity score and IOU overlap score to obtain a similarity score for every object in the tentative track and generate a sequence of similarity scores corresponding to all the objects of the tentative track $j$.
Note that we do not update the representative information of given track $i$ with any of the objects of tentative track $j$, unlike experiment 1.

To compute $z_{k+1}^{ij}(x_k)$, we calculate the average of all the previously calculated similarity scores.
We present a modified $F$ for this ablation study and refer to it as $F_3$. 
It is given by
\begin{equation}\label{eq:f3_score}
   F_3\big(T^i(x_{k}), T^j(\Bar{x}_{\ell}^{k+1})\big)=\frac{1}{ \ell}\sum_{b=k+1}^{k+\ell+1}\min\big\{\text{IOU}(\tilde{p}_{b}^{ij},\bar{p}^j_b), \textsc{CS}(\Tilde{v}_k^i,\bar{v}^j_b)\big\},
\end{equation}
where $\Tilde{p}^{ij}_{k+1},\dots,\Tilde{p}^{ij}_{k+\ell},,\Tilde{p}^{ij}_{k+\ell+1}$ are computed via \eqref{eq:track_p} and \eqref{eq:tentative_p}, as in experiment 1. 
The information $\Tilde{v}_k^i$ represents the representative visual information of a given track $i$, which can be viewed as a moving average of the sequence $v_0^i,v_1^i,\dots,v_k^i$.
This is different from experiment 2 and experiment 4 where we compare individual visual information of previously matched objects of a given track $i$.
Therefore, given a given track $i$ and tentative track $j$, this experiment tests how important it is to compare the visual information of previously matched objects of given track $i$ and objects of tentative track $j$ directly. 

The video-wise results for BoT-SORT and ADPTrack with BoT-SORT as the base heuristic are presented in Tables \ref{tab:videowise-botsort} and \ref{tab:experiment3}, respectively. The overall results are mentioned in Table \ref{tab:comparison-val-ablationstudies}. 
We see a significant improvement in the IDF1 scores of the following videos: MOT17-09 (7.313\%), MOT17-10 (4.76\%), MOT17-02 (1.1\%), MOT17-11(1.89\%). We see a small improvement for some videos (MOT17-04 and MOT17-13) and a drop in the IDF1 metric for the other videos (MOT17-05). 
Overall, we see an improvement of 1.16\% in the IDF1 metric, 0.4\% in the MOTA metrics, and 0.8\% in the HOTA metrics accompanied by a reduction in IDSW, FP, and FN metrics.  

\begin{table}[H]
\caption{Video-wise scores of ADPTrack with BoT-SORT as a base-heuristic as per experiment 3 over the validation dataset.}
\label{tab:experiment3}
    \centering
    \begin{tabular}{|l|c|c|c|c|c|c|c|}
        \hline
        Video Name & IDF1($\uparrow$) & MOTA($\uparrow$) & HOTA($\uparrow$) & FP($\downarrow$) & FN($\downarrow$) & IDSW($\downarrow$) & Frag($\downarrow$) \\ \hline
        MOT17-02 & 63.844 & 61.326 & 54.267 & 961 & 2793 & 67 & 95 \\ 
        MOT17-04 & 91.887 & 91.033 & 79.969 & 908 & 1240 & 20 & 33 \\ 
        MOT17-05 & 80.743 & 82.246 & 65.415 & 214 & 364 & 18 & 21 \\ 
        MOT17-09 & 84.349 & 88.017 & 70.839 & 22 & 317 & 6 & 5 \\ 
        MOT17-10 & 84.515 & 74.911 & 63.025 & 259 & 1217 & 10 & 55 \\ 
        MOT17-11 & 84.623 & 71.618 & 71.861 & 612 & 665 & 5 & 13 \\ 
        MOT17-13 & 90.193 & 83.587 & 71.156 & 85 & 432 & 1 & 5 \\ 
        \hline
    \end{tabular}
\end{table}

\subsubsection{Parameter Study for Experiment 3}
The number of frames $\ell$ for near-online simulation and the tuning parameter $\alpha$ used in \eqref{eq:c_def} are varied in two ablation studies and the experimentation results are presented in Tables \ref{tab:experiment3-ablationstudy} and \ref{tab:experiment3-ablationstudy2}, respectively.
We observe that the performance of the ADPTrack increases as we increase both the number of frames and then plateaus as we go further.

\begin{table}[H]
\centering
\caption{An ablation study over the number of subsequent frames $\ell$; ADPTrack with BoT-SORT as a base-heuristic as per experiment 3 over the validation dataset;}
\label{tab:experiment3-ablationstudy}
\begin{tabular}{|l|c|c|c|c|c|c|c|}
\hline
$\ell$ & IDF1($\uparrow$) & MOTA($\uparrow$) & HOTA($\uparrow$) & FP($\downarrow$) & FN($\downarrow$) & IDSW($\downarrow$) & Frag($\downarrow$) \\
\hline
1 & 84.004 & 81.047 & 71.074 & 9204 & 21066 & 372 & 663  \\
2 & 83.925 & 80.982 & 71.027 & 9276 & 21087 & 384 & 675  \\
3 & 83.925 & 80.982 & 71.027 & 9276 & 21087 & 384 & 675  \\
4 & 83.925 & 80.982 & 71.027 & 9276 & 21087 & 384 & 675  \\
5 & 83.929 & 80.995 & 71.03 & 9258 & 21084 & 384 & 675  \\
6 & 83.945 & 81.035 & 71.08 & 9180 & 21093 & 387 & 681  \\
7 & 84.309 & 81.024 & 71.303 & 9195 & 21096 & 387 & 684  \\
8 & 84.309 & 81.024 & 71.303 & 9195 & 21096 & 387 & 684  \\
9 & 84.309 & 81.024 & 71.303 & 9195 & 21096 & 387 & 684  \\
10 & 84.309 & 81.024 & 71.303 & 9195 & 21096 & 387 & 684  \\
11 & 84.444 & 81.043 & 71.403 & 9183 & 21084 & 381 & 681  \\
12 & 84.444 & 81.043 & 71.403 & 9183 & 21084 & 381 & 681  \\
13 & 84.444 & 81.043 & 71.403 & 9183 & 21084 & 381 & 681  \\
14 & 84.444 & 81.043 & 71.403 & 9183 & 21084 & 381 & 681  \\
15 & 84.413 & 81.021 & 71.385 & 9183 & 21120 & 381 & 684  \\
\hline
\end{tabular}
\end{table}

\begin{table}[H]
\centering
\caption{An ablation study over the tuning parameter $\alpha$. ADPTrack with BoT-SORT as a base heuristic as per experiment 3 over the validation dataset; $\ell$: number of subsequent frames for near-online simulation.}
\label{tab:experiment3-ablationstudy2}
\begin{tabular}{|l|l|c|c|c|c|c|c|c|}
\hline
$\ell$ & $\alpha$ & IDF1($\uparrow$) & MOTA($\uparrow$) & HOTA($\uparrow$) & FP($\downarrow$) & FN($\downarrow$) & IDSW($\downarrow$) & Frag($\downarrow$) \\
\hline
9 & 0.05 & 84.214 & 80.92 & 71.145 & 9285 & 21168 & 393 & 669  \\
9 & 0.1 & 84.309 & 81.024 & 71.303 & 9195 & 21096 & 387 & 684  \\
9 & 0.15 & 83.837 & 81.05 & 71.034 & 9231 & 21015 & 390 & 687  \\
9 & 0.2 & 83.501 & 81.024 & 70.891 & 9120 & 21138 & 420 & 702  \\
10 & 0.05 & 84.214 & 80.92 & 71.144 & 9285 & 21168 & 393 & 669  \\
10 & 0.1 & 84.309 & 81.024 & 71.303 & 9195 & 21096 & 387 & 684  \\
10 & 0.15 & 83.837 & 81.05 & 71.034 & 9231 & 21015 & 390 & 687  \\
10 & 0.2 & 83.533 & 81.06 & 70.942 & 9072 & 21132 & 417 & 702  \\
11 & 0.05 & 84.214 & 80.92 & 71.144 & 9285 & 21168 & 393 & 669  \\
11 & 0.1 & 84.444 & 81.043 & 71.403 & 9183 & 21084 & 381 & 681  \\
11 & 0.15 & 83.839 & 81.054 & 71.036 & 9228 & 21012 & 390 & 687  \\
11 & 0.2 & 83.533 & 81.06 & 70.942 & 9072 & 21132 & 417 & 702  \\
12 & 0.05 & 84.214 & 80.92 & 71.144 & 9285 & 21168 & 393 & 669  \\
12 & 0.1 & 84.444 & 81.043 & 71.403 & 9183 & 21084 & 381 & 681  \\
12 & 0.15 & 83.837 & 81.041 & 71.037 & 9219 & 21042 & 390 & 687  \\
12 & 0.2 & 83.533 & 81.06 & 70.942 & 9072 & 21132 & 417 & 702  \\
13 & 0.05 & 84.214 & 80.92 & 71.144 & 9285 & 21168 & 393 & 669  \\
13 & 0.1 & 84.444 & 81.043 & 71.403 & 9183 & 21084 & 381 & 681  \\
13 & 0.15 & 83.837 & 81.041 & 71.037 & 9219 & 21042 & 390 & 687  \\
13 & 0.2 & 83.533 & 81.06 & 70.942 & 9072 & 21132 & 417 & 702  \\
\hline
\end{tabular}
\end{table}

\subsubsection{Experiment 4}
We refer to the combination of experiments 1,2 and 3 presented in Algorithm \ref{alg:adptrack} as Experiment 4.
We present the overall scores of all the experiments in Table~\ref{tab:comparison-val-ablationstudies}. 

We observe that objects that are either mostly occluded or distant from the given track may not need to be considered in the evaluation of $F$ defined in \eqref{eq:prev_future_score}.
To address this, we introduce a heuristic that will discard certain objects from consideration for each tentative track $j$ with respect to a given track $i$.
We shortlist objects from the tentative track $T^j(\bar{x}_{\ell}^{k+1})$ and refer to them as candidates.
These candidate objects will be used to calculate $z_{k+1}^{ij}$ in \eqref{eq:prev_future_score}.
An object in a tentative track becomes a candidate for computations in \eqref{eq:prev_future_score} if the following holds: 
(i) The IOU overlap score of the object with the given track's estimated next state is sufficiently large, and 
(ii) The visual similarity between the object and the given track's representative visual information is sufficiently high. 
The number of candidate objects may be different for each tentative track $T^j(\bar{x}_{\ell}^{k+1})$ and given track $i$, and this value should replace $\ell$ in the denominator in \eqref{eq:prev_future_score} to produce a legitimate average value.

We observed that our algorithm may experience difficulty in the presence of objects that are continuously occluded, as their visual information does not capture consistent visual information of a single object, rather they may be a noisy mix of several intercepting objects. 
Therefore we propose a crowd-detection heuristic where we check if the objects have been occluded for an extended period with respect to their overall life and decide whether to use the proposed algorithm for performing associations for those tracks.
In cases such as these, the motion model alone may itself be a better estimator of the association scores.

This framework renders the computation of similarity scores between the given tracks and tentative tracks suitable for parallelization.
We have adopted multi-processing where a particular process computes the similarity score between a given track and a tentative track.
This parallelization can be further extended to object-to-object similarity score computations.

\begin{table}[H]
    \centering
    \caption{Comparison of proposed algorithms over the validation dataset.}
    \label{tab:comparison-val-ablationstudies}
    \begin{tabular}{|l|c|c|c|c|c|c|c|}
        \hline
        Algorithm & IDF1($\uparrow$) & MOTA($\uparrow$) & HOTA($\uparrow$) & FP($\downarrow$) & FN($\downarrow$) & IDSW($\downarrow$) & Frag($\downarrow$) \\ \hline
        BoT-SORT & 83.277 & 80.718 & 70.607 & 9312 & 21432 & 429 & 675\\
        Experiment 2 & 84.436 & 80.891 & 71.219 & 9378 & 21144 & 372 & 660  \\
        Experiment 3 & 84.444 & 81.043 & 71.403 & 9183 & 21084 & 381 & 681  \\
        Experiment 4 & 85.355 & 81.011 & 71.749 & 9252 & 21090 & 357 & 657 \\
        \hline
    \end{tabular}
\end{table}

\subsubsection{Parameter Study for Experiment 4}
In this section, we present ablation studies for various parameters present in Experiment 4 (Algorithm \ref{alg:adptrack}). 
Table~\ref{tab:experiment4-ablationstudy} presents an ablation study performed over the number of subsequent frames considered. 
The ablation study over the tuning parameter ($\alpha$) parameter is mentioned in Tables~\ref{tab:experiment4-weight1} and \ref{tab:experiment4-weight2}.
For generalization and application to the test dataset, we select the tuning parameter as 0.25.
We consider 15 and 5 for the number of subsequent frames ($\ell$) and the number of previously matched objects ($s$) from given tracks, respectively. 

\begin{table}[H]
\centering
\caption{An ablation study over the number of subsequent frames $\ell$ as per experiment 4.}
\label{tab:experiment4-ablationstudy}
\begin{tabular}{|l|c|c|c|c|c|c|c|}
\hline
$\ell$ & IDF1($\uparrow$) & MOTA($\uparrow$) & HOTA($\uparrow$) & FP($\downarrow$) & FN($\downarrow$) & IDSW($\downarrow$) & Frag($\downarrow$) \\
\hline
1 & 84.768 & 80.842 & 71.286 & 9453 & 21135 & 384 & 672  \\
2 & 84.38 & 80.976 & 71.125 & 9396 & 20976 & 384 & 663  \\
3 & 84.599 & 80.987 & 71.237 & 9363 & 21012 & 363 & 657  \\
4 & 84.641 & 80.974 & 71.31 & 9330 & 21054 & 375 & 672  \\
5 & 84.451 & 80.941 & 71.285 & 9342 & 21096 & 375 & 675  \\
6 & 84.904 & 80.946 & 71.434 & 9315 & 21123 & 366 & 660  \\
7 & 85.289 & 80.941 & 71.631 & 9300 & 21150 & 363 & 657  \\
8 & 85.147 & 80.995 & 71.721 & 9282 & 21090 & 354 & 654  \\
9 & 85.345 & 80.98 & 71.743 & 9294 & 21102 & 354 & 645  \\
10 & 85.148 & 81.037 & 71.721 & 9246 & 21057 & 354 & 660  \\
11 & 85.148 & 81.037 & 71.722 & 9246 & 21057 & 354 & 660  \\
12 & 85.341 & 81.045 & 71.746 & 9249 & 21042 & 354 & 651  \\
13 & 85.341 & 81.048 & 71.746 & 9249 & 21036 & 354 & 657  \\
14 & 85.355 & 81.011 & 71.749 & 9252 & 21090 & 357 & 657  \\
15 & 85.334 & 81.008 & 71.747 & 9255 & 21093 & 357 & 660  \\
\hline
\end{tabular}
\end{table}

\begin{table}[H]
\centering
\caption{An ablation study over the tuning parameter ($\alpha$) as per experiment 4. $\ell$: number of subsequent frames for near-online simulation. The value of $\ell$ varies between 6 and 10.}
\label{tab:experiment4-weight1}
\begin{tabular}{|l|l|c|c|c|c|c|c|c|}
\hline
$\ell$ & $\alpha$ & IDF1($\uparrow$) & MOTA($\uparrow$) & HOTA($\uparrow$) & FP($\downarrow$) & FN($\downarrow$) & IDSW($\downarrow$) & Frag($\downarrow$) \\
\hline
6 & 0.15 & 84.936 & 81.143 & 71.628 & 9141 & 21003 & 342 & 642  \\
6 & 0.2 & 84.858 & 81.058 & 71.477 & 9141 & 21126 & 357 & 651  \\
6 & 0.25 & 84.904 & 80.946 & 71.434 & 9315 & 21123 & 366 & 660  \\
6 & 0.3 & 84.565 & 80.841 & 71.207 & 9396 & 21195 & 384 & 675  \\
6 & 0.35 & 84.278 & 80.67 & 71.038 & 9474 & 21360 & 417 & 681  \\
7 & 0.15 & 84.937 & 81.139 & 71.629 & 9144 & 21009 & 339 & 642  \\
7 & 0.2 & 84.912 & 81.073 & 71.59 & 9159 & 21096 & 345 & 645  \\
7 & 0.25 & 85.289 & 80.941 & 71.631 & 9300 & 21150 & 363 & 657  \\
7 & 0.3 & 84.51 & 80.848 & 71.193 & 9387 & 21189 & 387 & 681  \\
7 & 0.35 & 84.475 & 80.781 & 71.125 & 9351 & 21315 & 405 & 678  \\
8 & 0.15 & 84.937 & 81.139 & 71.629 & 9144 & 21009 & 339 & 642  \\
8 & 0.2 & 84.924 & 81.119 & 71.594 & 9150 & 21030 & 345 & 651  \\
8 & 0.25 & 85.147 & 80.995 & 71.721 & 9282 & 21090 & 354 & 654  \\
8 & 0.3 & 84.528 & 80.842 & 71.192 & 9360 & 21228 & 384 & 678  \\
8 & 0.35 & 84.488 & 80.813 & 71.131 & 9324 & 21294 & 402 & 675  \\
9 & 0.15 & 84.937 & 81.139 & 71.629 & 9144 & 21009 & 339 & 642  \\
9 & 0.2 & 84.924 & 81.11 & 71.627 & 9177 & 21021 & 342 & 651  \\
9 & 0.25 & 85.345 & 80.98 & 71.743 & 9294 & 21102 & 354 & 645  \\
9 & 0.3 & 84.621 & 80.922 & 71.383 & 9288 & 21189 & 366 & 666  \\
9 & 0.35 & 84.972 & 80.889 & 71.489 & 9207 & 21300 & 390 & 669  \\
10 & 0.15 & 84.935 & 81.141 & 71.629 & 9147 & 21003 & 339 & 645  \\
10 & 0.2 & 84.928 & 81.121 & 71.629 & 9153 & 21027 & 342 & 651  \\
10 & 0.25 & 85.148 & 81.037 & 71.721 & 9246 & 21057 & 354 & 660  \\
10 & 0.3 & 84.597 & 80.941 & 71.373 & 9252 & 21192 & 369 & 663  \\
10 & 0.35 & 85.099 & 80.837 & 71.556 & 9249 & 21327 & 405 & 681  \\
\hline
\end{tabular}
\end{table}

\begin{table}[H]
\centering
\caption{Experiment 4: An ablation study over the tuning parameter ($\alpha$) as per experiment 4. $\ell$: number of subsequent frames for near-online simulation. The value of $\ell$ varies between 11 and 15.}
\label{tab:experiment4-weight2}
\begin{tabular}{|l|l|c|c|c|c|c|c|c|}
\hline
$\ell$ & $\alpha$ & IDF1($\uparrow$) & MOTA($\uparrow$) & HOTA($\uparrow$) & FP($\downarrow$) & FN($\downarrow$) & IDSW($\downarrow$) & Frag($\downarrow$) \\
\hline
11 & 0.15 & 84.959 & 81.102 & 71.591 & 9135 & 21072 & 345 & 645  \\
11 & 0.2 & 84.922 & 81.119 & 71.593 & 9153 & 21027 & 345 & 651  \\
11 & 0.25 & 85.148 & 81.037 & 71.722 & 9246 & 21057 & 354 & 660  \\
11 & 0.3 & 84.595 & 80.941 & 71.372 & 9255 & 21186 & 372 & 663  \\
11 & 0.35 & 85.044 & 80.859 & 71.501 & 9246 & 21297 & 402 & 675  \\
12 & 0.15 & 84.943 & 81.119 & 71.631 & 9150 & 21036 & 339 & 642  \\
12 & 0.2 & 84.922 & 81.119 & 71.592 & 9153 & 21027 & 345 & 651  \\
12 & 0.25 & 85.341 & 81.045 & 71.746 & 9249 & 21042 & 354 & 651  \\
12 & 0.3 & 84.849 & 80.915 & 71.502 & 9336 & 21159 & 360 & 666  \\
12 & 0.35 & 85.022 & 80.816 & 71.486 & 9318 & 21291 & 405 & 678  \\
13 & 0.15 & 84.943 & 81.119 & 71.631 & 9150 & 21036 & 339 & 642  \\
13 & 0.2 & 84.922 & 81.119 & 71.592 & 9153 & 21027 & 345 & 651  \\
13 & 0.25 & 85.341 & 81.048 & 71.746 & 9249 & 21036 & 354 & 657  \\
13 & 0.3 & 84.813 & 80.82 & 71.473 & 9417 & 21228 & 363 & 666  \\
13 & 0.35 & 84.671 & 80.811 & 71.339 & 9345 & 21279 & 399 & 678  \\
14 & 0.15 & 84.937 & 81.117 & 71.594 & 9150 & 21036 & 342 & 642  \\
14 & 0.2 & 84.922 & 81.119 & 71.592 & 9153 & 21027 & 345 & 651  \\
14 & 0.25 & 85.355 & 81.011 & 71.749 & 9252 & 21090 & 357 & 657  \\
14 & 0.3 & 85.367 & 80.842 & 71.706 & 9381 & 21225 & 366 & 660  \\
14 & 0.35 & 85.059 & 80.811 & 71.524 & 9312 & 21306 & 405 & 678  \\
15 & 0.15 & 84.938 & 81.119 & 71.595 & 9147 & 21036 & 342 & 642  \\
15 & 0.2 & 84.918 & 81.11 & 71.59 & 9162 & 21033 & 345 & 654  \\
15 & 0.25 & 85.334 & 81.008 & 71.747 & 9255 & 21093 & 357 & 660  \\
15 & 0.3 & 85.367 & 80.842 & 71.705 & 9381 & 21225 & 366 & 660  \\
15 & 0.35 & 84.636 & 80.815 & 71.332 & 9333 & 21279 & 405 & 672  \\
\hline
\end{tabular}
\end{table}


\end{document}